\documentclass[final]{elsarticle}

\usepackage{amssymb}
\usepackage{amsmath}

\usepackage{subfig}
\usepackage{graphicx}
\usepackage[dvipsnames]{xcolor}
\usepackage[colorlinks=true,citecolor=blue]{hyperref}
\usepackage{caption}
\usepackage{multirow}
\usepackage{pdftexcmds}
\usepackage{wrapfig}
\usepackage{color}

\journal{}

\begin{document}

\begin{frontmatter}

\title{An efficient heuristic for geometric analysis of cell deformations}

\author[1]{Yaima Paz Soto\corref{cor1}}
\ead{ysoto@cug.co.cu}
\cortext[cor1]{}

\author[2]{Silena Herold Garcia}
\ead{silena@uo.edu.cu}

\author[3]{Ximo Gual-Arnau}
\ead{gual@uji.es}

\author[4,6,7,8]{Antoni Jaume-i-Capó}
\ead{antoni.jaume@uib.es}

\author[5,6,7,8]{Manuel González-Hidalgo}
\ead{manuel.gonzalez@uib.es}

\affiliation[1]{organization={Department of Informatics, University of Guantánamo},
            city={Guantánamo},
           	country={Cuba}}

\affiliation[2]{organization={Department of Computer Science, University of Oriente},
	city={Santiago de Cuba},
	country={Cuba}}

\affiliation[3]{organization={Departament de Matemàtiques, Institute of New Imaging Technologies, Universitat Jaume I},
	city={Castelló},
	country={Spain}}

\affiliation[4]{organization={UGiVIA Research Group, University of the Balearic Islands, Dpt. of Mathematics and Computer Science},
	addressline={Crta. Valldemossa, Km 7.5},
	city={Palma},
	postcode={E-07122},
	country={Spain}}
	
\affiliation[5]{organization={SCOPIA Research Group, University of the Balearic Islands, Dpt. of Mathematics and Computer Science},
	addressline={Crta. Valldemossa, Km 7.5},
	city={Palma},
	postcode={E-07122},
	country={Spain}}
	
\affiliation[6]{organization={Health Research Institute of the Balearic Islands (IdISBa)},
	city={Palma},
	postcode={E-07122},
	country={Spain}}

\affiliation[7]{organization={Laboratory for Artificial Intelligence Applications at UIB (LAIA@UIB)},
city={Palma},
postcode={E-07122},
country={Spain}}

\affiliation[8]{organization={Artificial Intelligence Research Institute of the Balearic Islands (IAIB)},
city={Palma},
postcode={E-07122},
country={Spain}}

\begin{abstract}

{Sickle cell disease causes erythrocytes to become sickle-shaped, affecting their movement in the bloodstream and reducing oxygen delivery. It has a high global prevalence and places a significant burden on healthcare systems, especially in resource-limited regions. Automated classification of sickle cells in blood images is crucial, allowing the specialist to reduce the effort required and avoid errors when quantifying the deformed cells and assessing the severity of a crisis. Recent studies have proposed various erythrocyte representation and classification methods \cite{jennifer23}. Since classification depends solely on cell shape, a suitable approach models erythrocytes as closed planar curves in shape space \cite{epifanio:2020}. This approach employs elastic distances between shapes, which are invariant under rotations, translations, scaling, and reparameterizations, ensuring consistent distance measurements regardless of the curves’ position, starting point, or traversal speed. While previous methods exploiting shape space distances had achieved high accuracy, we refined the model by considering the geometric characteristics of healthy and sickled erythrocytes. Our method proposes (1) to employ a fixed parameterization based on the major axis of each cell to compute distances and (2) to align each cell with two templates using this parameterization before computing distances. Aligning shapes to templates before distance computation, a concept successfully applied in areas such as molecular dynamics \cite{richmond04}, and using a fixed parameterization, instead of minimizing distances across all possible parameterizations, simplifies calculations. This strategy achieves 96.03\% accuracy rate in both supervised classification and unsupervised clustering. Our method ensures efficient erythrocyte classification, maintaining or improving accuracy over shape space models while significantly reducing computational costs.}
\end{abstract}




\end{frontmatter}


\section{Introduction}
\label{intro}




Automatic shape analysis has recently gained increasing interest in the medical community due to its potential to precisely identify morphological changes between different populations of structures in images. When combined with statistical analysis, this approach can provide measurements of structural change at corresponding locations.

{Sickle cell anemia (also known as drepanocytic anemia) leads to the deformation of red blood cells (erythrocytes), which lose their typical round shape and take on a crescent or sickle shape. This morphological change increases the risk of various complications, including vaso-occlusive events and painful crises. The clinical assessment of a patient’s condition typically involves microscopic examination of the different types of erythrocytes, relying on their variable morphology. Accurate classification of these red blood cells is crucial for diagnosis, as it helps determine the severity of the crisis. However, quantifying the number of sickle cells presents challenges, including the variability of criteria used by different specialists, the lack of standardized evaluation methods, and the labor-intensive, error-prone nature of the process. This requires specialized personnel and a considerable amount of time, and delays or inaccuracies in quantification can affect patient outcomes.}

{A promising solution for these challenges is tautomatically analyzing of erythrocyte shapes in peripheral blood smear images. In recent years, various techniques for red blood cell classification have been developed, demonstrating that automated methods are not only faster and more scalable for large patient populations, but also more reliable than manual inspection for cellular classification. Before discussing the motivation behind this work and the specific challenges it aims to address, we will first review some of these existing methods to provide context for the problem.}

The first approach to be used, and one of the simplest, is based on elementary shape descriptors (e.g., circularity and ellipticity). These descriptors have been employed in studies such as \cite{claib:2011} and \cite{mojt:2013}, but their simplicity has limited their efficiency in classification tasks. However, more recent works, such as \cite{rodrigues2016morphological}, consider a broader set of descriptors, which, when combined with machine learning classification algorithms, yield higher accuracy.
Fourier series-based descriptors have been widely used in the analysis of erythrocyte shape, often combined with other characteristics like geometric features, as seen in \cite{sherna:2013}, or with centroid-contour distance and aspect ratio, as in \cite{Deb:2014}. Various transforms for detecting circular objects, such as the Hough transform and the circlet method, have been applied in \cite{mazalan:2013} and \cite{sarrafzadeh:2015}. Additionally, mathematical morphology has been used for shape analysis in \cite{maji:2015} and \cite{chandrasiri:2014}.
Template matching is another approach used in several studies. For example, in \cite{bronkorsta:2000}, a parametric deformable template model was employed to recover the shape of red blood cells. Similarly, in \cite{frejlichowski:2010} and \cite{bergen:2008}, known erythrocyte templates were used for classification. Artificial neural networks have also been applied for red blood cell classification in studies such as \cite{tomari:2015}, \cite{elsalamony:2016}, \cite{lee:2014}, and \cite{paz:2020}. Another commonly used classifier is the Support Vector Machine (SVM), which was utilized in \cite{akrimi:2014} and in combination with neural networks and k-Nearest Neighbors ($k$-NN) in \cite{lotfi:2015} to study erythrocyte abnormalities.
Ellipse fitting methods have been employed extensively to detect elliptical-shaped cells \cite{goncalves:2011, liao:2016}, and to address the problem of overlapping cells, as discussed in \cite{gonzalez:2015}. In \cite{delgado:2016}, Hidden Markov Models were used to model cell contours. Integral geometry-based functions were applied in  \cite{gual:2013}, and \cite{gualetal:2015} to extract shape features and obtain suitable representations of erythrocytes. Some methods have also been developed to handle different imaging conditions, including techniques for extracting 3D cell shapes, as shown in \cite{simionato:2021} and applied in \cite{darras:2021}. However, these techniques typically require more advanced and expensive setups that are not yet suitable for point-of-care testing devices.
These studies represent just a subset of the most recent advancements in erythrocyte morphological analysis.

 {However, when classifying planar shapes, a historically used geometric characterization involves representing these shapes through the curves that define their continuous boundaries. In this approach, planar shapes, such as red blood cells in our case, are treated as simple closed plane curves that form the space of planar shapes. Within this space, distances between shapes can be defined, which allows for the computation of, for example, mean shapes and interpolation between shapes. Each planar shape is then characterized by a parameterized curve, and the set of all planar shapes (the space of planar shapes) is a Riemannian manifold with an appropriate distance metric, as discussed in \cite{younes:2008}.
Shape analysis of curves in the shape space has been applied to various data science problems, particularly classification tasks, as shown in studies such as \cite{kurtek12} and \cite{laga12}. It has also been used specifically for the classification of red blood cells, as demonstrated in \cite{gual:2015} and \cite{epifanio:2020}.}

 {As we will see in Section \ref{section:methods}, how the space of planar shapes is defined ensures that the distance between shapes (curves) remains invariant under both scaling transformations and motions applied to these shapes. In this context, curves in the shape space are parametrized curves, defined as functions that include information on how the curve is traversed (e.g., starting point and velocity). However, the curves of interest for representing planar shapes are geometric curves, which do not depend on parameterization. Therefore, to compute the distance between shapes, one must minimize the distances across all possible parameterizations that define the shapes (geometric curves). This approach, while yielding excellent classification results, is computationally expensive.}

 {In \cite{gual:2015, epifanio:2020}, the representation of red blood cells as elements in the  shape space is considered, utilizing two distinct Sobolev-type distances. Both approaches deliver excellent classification results. However, in one case \cite{gual:2015}, only arc-length parameterizations of the curve are considered. This simplifies the computation of distances between shapes but negatively impacts classification accuracy. In \cite{epifanio:2020}, a distance is used that minimizes overall parameterizations (elastic distance), resulting in significantly improved classification performance.}

 {The novelty of the technique presented in this work is that it leverages the specific geometry of healthy red blood cells (resembling a circle) and sickle cells (resembling a sickle shape) to fix a unique parameterization based on the major axis of each cell. Distances are then calculated using this fixed parameterization. Thus, by constraining the curves in this way, the computational methods required to calculate distances are simplified, as not all parameterizations are considered.}

 {Additionally, each cell is aligned using this parameterization to two templates (a circle and an ellipse), and red blood cell classification is performed by calculating the distances of each shape, with the fixed parameterization, to these templates. In Section \ref{Result_Disc}, we conduct various classification experiments based on a dataset of red blood cells using our technique, and comparing our results with those of \cite{gual:2015, epifanio:2020}, where minimization over all parameterizations is used to compute distances. As we will show, the computational cost is significantly reduced, while the classification accuracy improves over \cite{gual:2015} and remains on par with \cite{epifanio:2020}, where an elastic distance was used.}
 
 {The excellent performance of the method ensures its potential to support early diagnosis and personalized treatment plans, if it is implemented in tools designed to assist specialists in the clinical management of patients with sickle cell disease. This has broader implications for resource-limited regions, as it can enhance healthcare equity by improving access to accurate diagnoses. However, while our method produces good classification results and simplifies calculations, its performance is limited by its reliance on the geometric characteristics of the templates used, making it less effective for shapes that do not share these characteristics.}
 

  {The structure of this paper is as follows: In Section \ref{section:methods}, we provide an overview of the two shape spaces used in this study, each considering a different metric as described in \cite{gual:2015} and \cite{epifanio:2020}. We then introduce our proposed heuristic strategy by defining a fixed parameterization and explain how distances between shapes are computed using this approach.
Section \ref{Result_Disc} presents the image database utilized in this work and describes the four experiments conducted, along with the performance indicators used to evaluate and compare the results. These experiments include both supervised classification and unsupervised clustering, considering elastic metrics in the shape spaces, as well as distances based on our fixed parameterization and distances to two specific templates.
In Section \ref{discussion}, we display and analyze comparative tables that showcase the performance indicators for the various experiments. We also conduct a study demonstrating that our proposed method significantly reduces computational cost, and we compare our accuracy results with those of existing models using the same dataset. Also, in this section we present the application method to other shape analysis scenarios.
Finally, Section \ref{conclu} summarizes the main findings of the study, highlights the significance of our approach, and discusses the identified limitations. }

\section{Methods}
\label{section:methods}

{In this section, for the sake of completeness and clarity, we will complete and review some results from \cite{gual:2015} and \cite{younes:2008} (Section \ref{sec:s1Space}) and from \cite{epifanio:2020} and \cite{srivastava:2011} (Section \ref{sec:s2Space}). In addition, we will introduce a new proposal on the calculation of distances between curves with a fixed parameterization (Section \ref{subsec:fixed_paramet}).}

\subsection{The space of planar shapes $\mathcal{S}_1$: the Grassmann manifold model}
\label{sec:s1Space}

In this section, we review some results of \cite{gual:2015} and \cite{younes:2008}. Let $M$ be the set of closed curves that are smooth boundaries of planar shapes, i.e. 
$$M=\{\alpha \in C^{\infty} (\mathbb {S}^1, \mathbb {R}^2)\,\, :\,\, |\alpha'(t)|\neq 0,\,\,\forall t\in\mathbb {S}^1 \},$$
where $\mathbb {S}^1$ denotes the unit circle, {$\alpha'(t)$ stands for the parametric derivative of $\alpha$, and $|\alpha'(t)|$ is its euclidean norm}. $M$ is for the space of $C^{\infty}$-immersions $\alpha:[0,2\pi]\longrightarrow \mathbb {R}^2$ with $\alpha(0)=\alpha(2\pi)$.

Let the tangent space \( T_{\alpha}M \) at \( \alpha \) be the set of vector fields \( h \) on \( \alpha \) (\( h: \mathbb{S}^1 \longrightarrow \mathbb{R}^2 \)). If \( h, k \in T_{\alpha}M \), the Riemannian metric \( G_{\alpha} (h, k) \) is defined as in \cite{younes:2008} by

\begin{displaymath}
	G_{\alpha} (h,k) =\frac 1{l(\alpha)}\int_{S^1}\dot{h}(s) \bullet \dot{k}(s)\, ds,
\end{displaymath}
where $l(\alpha)$ is the length of $\alpha$, $\dot{h}(s)$ denotes derivative with respect to arc length, and $\dot{h}(s) \bullet \dot{k}(s)$ stands for the usual product in $\mathbb {R}^2$.

The set \( M \) modulo translations, scalings, rotations, and reparameterizations of the curve (\( \text{Diff}(\mathbb{S}^1) \)) is the space of planar shapes. Nevertheless, let us first study the {\it pre-shape} space, i.e., before the division by the group of diffeomorphisms \( \text{Diff}(\mathbb{S}^1) \) is carried out. We refer to the group generated by translations, rotations, and scalings as the group of similitudes ({\it sim}). Therefore, the {\it pre-shape} space \( M_d \) is defined as the quotient space

\begin{displaymath}
	M_d=M /{\it sim},
\end{displaymath}
and the restriction of the metric \( G_{\alpha} \) is associated with \( M_d \).

\bigskip

We denote by \( V \) the vector space of all \( C^{\infty} \) mappings \( f: \mathbb{S}^1 \longrightarrow \mathbb{R} \), with the norm
\[
||f||^2 = \int_0^{2\pi} (f(x))^2 \, dx.
\]

On the other hand, given a curve \( \alpha \in M \), we can obtain the functions \( e \) and \( f \), which correspond to the representative of \( \alpha \) in \( M_d \), from the following expressions:

\begin{equation}\label{eq131}
	e (t)=\sqrt{\frac{2 |\alpha' (t)|}{l(\alpha)}} \cos\left(\frac{\theta_{\alpha} (t)}{2}\right),\quad f (t)=\sqrt{\frac{2 |\alpha' (t)|}{l(\alpha)}} \sin\left(\frac{\theta_{\alpha} (t)}{2}\right),
\end{equation}
{where $\theta_{\alpha} (t)$ is the tangent angle function of $\alpha$ and $l(\alpha)$ is the length of $\alpha$.}

Let \( e, f \) be two functions in \( V \), and let us assume that our plane curves are curves in the complex plane \( \mathcal{C} \).  
We define the basic mapping by
\begin{equation}\label{eq5}
	\Phi: (e,f)\longmapsto \alpha (t)=\frac12\int_0^{t} (e(x)+{\rm i} f(x))^2dx.
\end{equation}

Let  $Gr(2,V)$ be the Grassmannian  of unoriented 2-dimensional subspaces of $V$ defined by an orthonormal pair $(e,f)\in V^2$ with $ ||e||^2+||f||^2=2$ (i.e. $l(\alpha)=1$);  
then, $\Phi$ defines an isometry between $M_d$ and a subset of $Gr(2,V)$.

In order to compute distances between any pair of shapes (closed planar curves) in the pre-shape space $M_d$, we will consider the basic mapping and the geodesic distance in the Grassmannian between the points $(e, f)$ defined by the two functions $e(t)$ and $f(t)$ that represent the shapes.

Then, the geodesic distance between \( \alpha = \Phi(e_1, f_1) \) and \( \beta = \Phi(e_2, f_2) \) is the distance between the two-dimensional subspaces \( W_1 \), generated by \( \{e_1, f_1\} \), and \( W_2 \), generated by \( \{e_2, f_2\} \).

The singular value decomposition of the orthogonal projection \( p \) of \( W_1 \) onto \( W_2 \) gives orthonormal bases \( \{\hat{e}_1, \hat{f}_1\} \) of \( W_1 \) and \( \{\hat{e}_2, \hat{f}_2\} \) of \( W_2 \) such that \( p(\hat{e}_1) = \lambda_1 \hat{e}_2 \) and \( p(\hat{f}_1) = \lambda_2 \hat{f}_2 \), with \( \hat{e}_1 \perp \hat{f}_2 \), \( \hat{f}_1 \perp \hat{e}_2 \), where \( 0 \leq \lambda_1, \lambda_2 \leq 1 \).

In fact, \( \lambda_1 \) and \( \lambda_2 \) are the singular values of the \( (2 \times 2) \)-matrix.

\begin{equation}\label{matrix}
	A= \left( \begin{array}{cc}
		<e_1, e_2> & <e_1, f_2>\\
		<f_1, e_2> & <f_1, f_2>\end{array} \right),
\end{equation}
{where $<\cdot\, ,\, \cdot >$ denotes the inner product in $V$}. 

If we write $\lambda_1= \cos \psi_1$, $\lambda_2= \cos \psi_2$ then $\psi_1, \psi_2$ are the Jordan angles, $0\le\psi_1, \psi_2\le\pi/2$, and the geodesic distance between
$\alpha =\Phi (e_1, f_1)$ and $\beta =\Phi (e_2, f_2)$ is given by:

\begin{equation} \label{distance_preshape}
	d(\alpha, \beta)=d(W_1, W_2)=\sqrt{\psi_1^2+\psi_2^2}.
\end{equation}

The main interest in \cite{gual:2015} is to compute distances between geometric curves, i.e., curves considered up to reparameterizations. The shape space is then defined as the quotient space

\begin{equation}\label{shape_space}
	\mathcal{S}_1=M_d/ \text{Diff}(\mathbb {S}^1).
\end{equation}

The closed planar curves considered correspond to cell boundaries, when a fixed orientation and approximately equally spaced discrete points are used. We suppose that the curves \( \alpha \in M_d \) have constant speed \( |\alpha'(t)| \) for all \( t \in [0, 2\pi) \), and \( \phi \in \text{Diff}^+(\mathbb{S}^1) \) (i.e., \( C^{\infty} \) orientation-preserving diffeomorphisms of \( \mathbb{S}^1 \)) with \( \phi(0) = 0 \). Then, if \( \alpha(t) \) is a curve in \( M_d \) with \( |\alpha'(t)| = K \), and \( \alpha \circ \phi \) is a reparameterization of the curve with \( |(\alpha \circ \phi)'(t)| = K \), we have that \( \phi \) is the identity, i.e., \( \phi(t) = t \) for all \( t \in \mathbb{S}^1 \). Therefore, the distance in the shape space \( \mathcal{S}_1 \) between two closed curves \( \alpha \) and \( \beta \), each of length 1, representing two shapes, is

\begin{equation}\label{minpa}
	d(\alpha, \beta)=\min_{\phi }d(\alpha, \widetilde{\beta\circ\phi}),
\end{equation} 
where $\phi\in \text{Diff}^+( \mathbb {S}^1)$ is given by $\phi (t) = t + t_0$, $t_0\in\mathbb{S}^1$ is a constant, and $ \widetilde{\beta\circ\phi}$ is the representative element of $ \beta\circ\phi$ in $M_d$.\\

We will analyze contours characterizing cell boundaries with a fixed orientation and discrete points approximately evenly spaced along these contours.

The distance and the geodesic between two shapes in \( \mathcal{S}_1 \) are computed as in \cite{gual:2015}. However, as deformations of curves respect the arc-length parameter, stretch elasticity is not incorporated into the model, and the resulting shape correspondences are sometimes far from optimal. In this first proposal, only bending is considered for the evolution from one curve to another, so the metric is not fully exploited, and the behavior is similar to that of non-elastic metrics. In the rest of the article, references to the space \( \mathcal{S}_1 \) are made with this particularity in mind.

\subsection{The space of planar shapes $\mathcal{S}_2$: the Square Root Velocity Function representation model}\label{sec:s2Space}

In this section, we review some results from \cite{epifanio:2020} and \cite{srivastava:2011}. In particular, we consider the Square Root Velocity Function (SRVF) representation of closed curves in \( \mathbb{R}^2 \), and we summarize the main results for the shape space \( \mathcal{S}_2 \) with the standard elastic metric.

Let \( \beta:[0,1]\longrightarrow \mathbb{R}^2 \) be a parameterized curve that is absolutely continuous on \([0,1]\). The Square Root Velocity (SRV) of \( \beta \) is defined as the function \( q:[0,1]\longrightarrow \mathbb{R}^2 \) given by \cite{srivastava:2011}:

\begin{equation} \label{srv1}
	q(t)=\frac{\beta'(t)}{\sqrt{|\beta' (t)|}}.
\end{equation}

For every \( q \in \mathbb{L}^2 ([0,1], \mathbb{R}^2) \), there exists a curve \( \beta \) (unique up to translation) such that the given \( q \) is the SRV function of that \( \beta \). In fact,
\begin{equation} \label{srv2}
	\beta(t)=\int_0^t q(s)|q(s)|\, d s.
\end{equation}

To remove the scaling variability, we scale all curves to have unit length.
If a curve $\beta$ is of length one, then $\int_0^1 |q(t)|^2 dt =1$.

Therefore, the SRV functions associated with these curves are elements of the unit hypersphere in the Hilbert manifold $\mathbb{L}^2 ([0,1],\mathbb {R}^2)$: 
\begin{equation} \label{srv3}
	{\cal C}^0 =\{q:[0,1]\longrightarrow \mathbb {R}^2\,\, : \,\, \int_0^1 |q(t)|^2 dt =1\},
\end{equation}

For studying shapes of closed curves, we must impose the additional condition that the curve starts and ends at the same point. Then, the space of fixed-length closed curves represented by their SRVF is:
\begin{equation} \label{srv4}
	{\cal C}^c =\{q:\mathbb {S}^1\longrightarrow \mathbb {R}^2\,\, : \,\, \int_{\mathbb {S}^1} |q(t)|^2 dt =1,\, \,\, \int_{\mathbb {S}^1} q(t)|q(t)| dt =0\}.
\end{equation}

To address the rotation and reparameterization of the curve \( \beta \) whose SRVF is \( q \), we recall that a rotation is an element of \( SO(2) \), the special orthogonal group of \( 2 \times 2 \) matrices, and a reparameterization is an element of \( \Gamma \), the set of orientation-preserving diffeomorphisms of \( \mathbb{S}^1 \). The actions of \( SO(2) \) and \( \Gamma \) on the SRVF of \( \beta \) are given by: $(O, q(t)) = Oq(t)$, where $O\in SO(2)$, and the SRV of the curve $\beta\circ \gamma$ is $q(\gamma(t))\sqrt{\gamma' (t)}$, where $\gamma\in\Gamma$.

The orbit of a function $q\in{\cal C}^c$ is (see $C^c$ in Eq. \ref{srv4})
\begin{equation} \label{orbit}
	[q] =\{O(q,\gamma)= O(q(\gamma(t)))\sqrt{\gamma' (t)} \,\, :\,\, (\gamma ,O)\in \Gamma\times SO(2)\}.
\end{equation}

If we consider the metric in $\mathbb{L}^2$ given by the usual inner product
\begin{equation} \label{prod}
	\langle v_1, v_2\rangle_{\mathbb{L}^2} =\int_0^1 \langle v_1 (t), v_2 (t)\rangle\, dt,
\end{equation}
the feature space of interest is:

\begin{equation} \label{s2}
	{\cal S}_2=\{[q]\,\, : \,\, q\in{\cal C}^c\},
\end{equation} and the distance in ${\cal S}_2$ is:

\begin{equation} \label{dist_S2}
	d_s ([q_1], [q_2])= \inf_{O\in SO(2), \gamma\in\Gamma} d_c(q_1, O(q_2,\gamma)),
\end{equation} where $d_c$ denotes the distance in the hypersphere ${\cal C}^c$. The computation of the distances and geodesics in the shape space ${\cal S}_2$ was detailed by \cite{srivastava:2011} and \cite{joshi:2007}. 

{In both spaces \( \mathcal{S}_1 \) and \( \mathcal{S}_2 \), the performance of the classification procedure has also been studied considering the distance to the circle and ellipse templates, taking into account the similarity that the objects of interest have with them. It has been concluded that this variant can be used to reduce the computational cost of the classification process \cite{gual:2015,epifanio:2020}. With this in mind, classification experiments using circle and ellipse templates are carried out in this paper to test the performance of the method in this case.}

\subsection{Distances between curves with fixed parameterization}\label{subsec:fixed_paramet}

In the two shape spaces analyzed, the distance between two shapes is obtained by computing the distances between the first and all the reparameterizations of the second, and determining the minimum value in this iterative and time consuming process (see Eq.(\ref{minpa}) and Eq. (\ref{dist_S2})). 

{Our proposal takes into account the shape of sickle cells, which is the most important class for analysis in peripheral blood samples from patients with sickle cell anemia. We propose to establish the parameterization of shapes considering an equidistant distance between the contour points and to align the shapes considering their major axis and the $x$ axis. For better visualization of the method, in the experiments the major axis of the shapes is oriented according to the $x$ coordinate axis. Given that the calculations assume the centroid is positioned at the origin, the starting point for the contour parameterization corresponds to the point on the contour with the largest positive $x-$coordinate}. When performing classification using distances to templates, the choice of any of the two points that define the major axis does not affect the result of this classification. 

\begin{figure}[!ht]
	\begin{center}
	\subfloat[][d=0.7108]{\includegraphics[width=5cm]{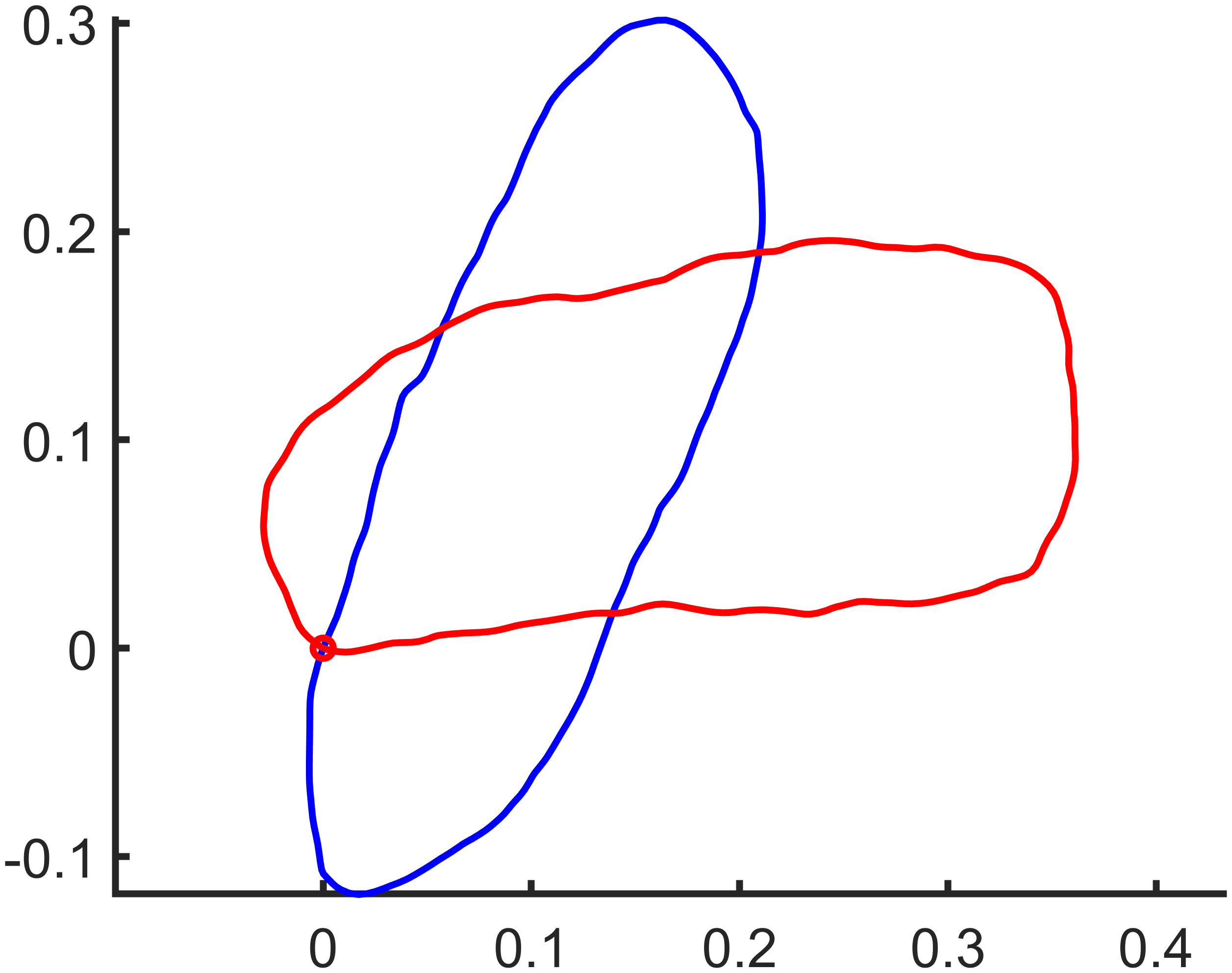}\label{fig:four reparameterizations1}}\quad
	\subfloat[][d=0.3586]{\includegraphics[width=5cm]{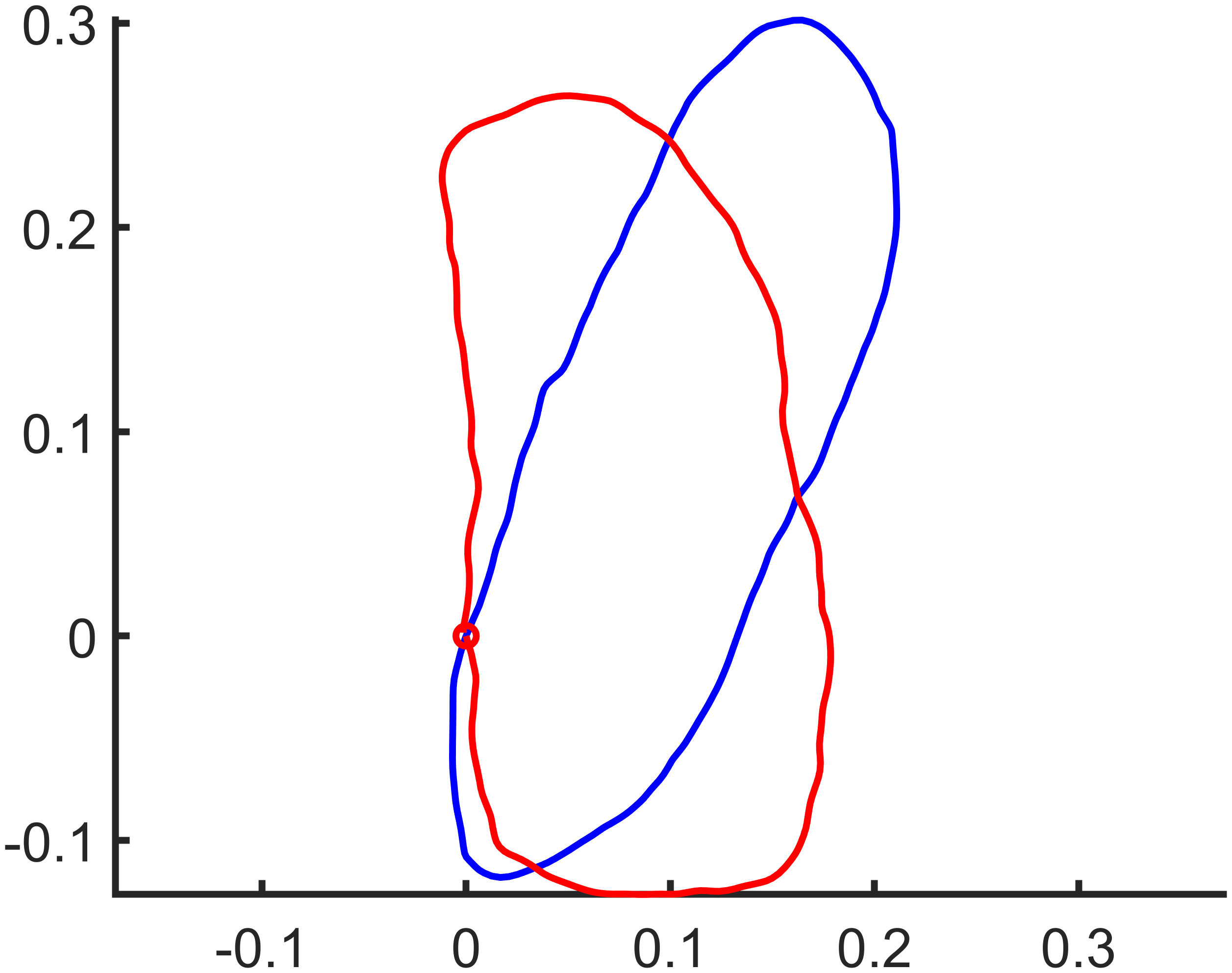}\label{fig:four reparameterizations2}}\\
	\subfloat[][d=0.6890]{\includegraphics[width=5cm]{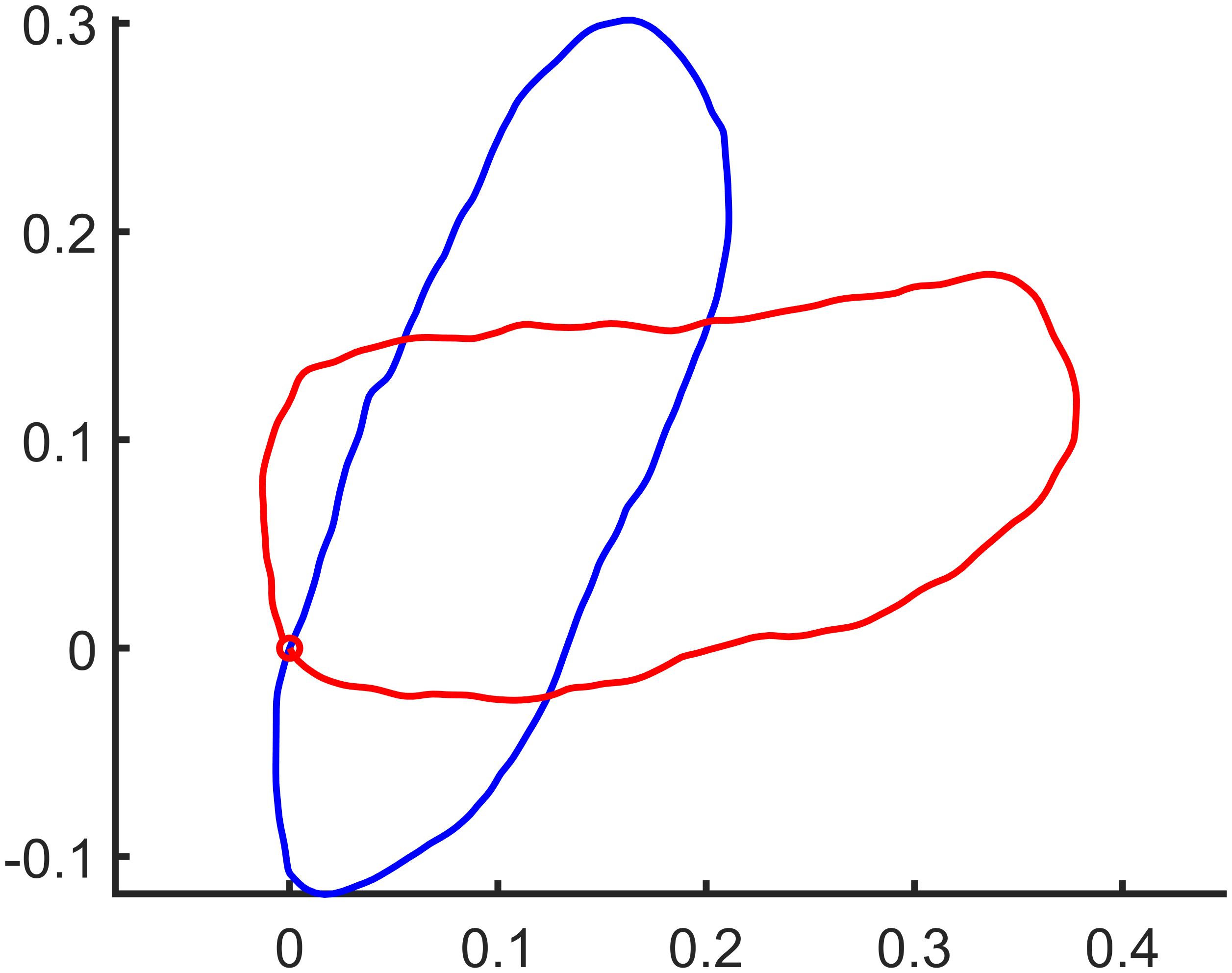}\label{fig:four reparameterizations3}}\quad
	\subfloat[][d=0.3765]{\includegraphics[width=5cm]{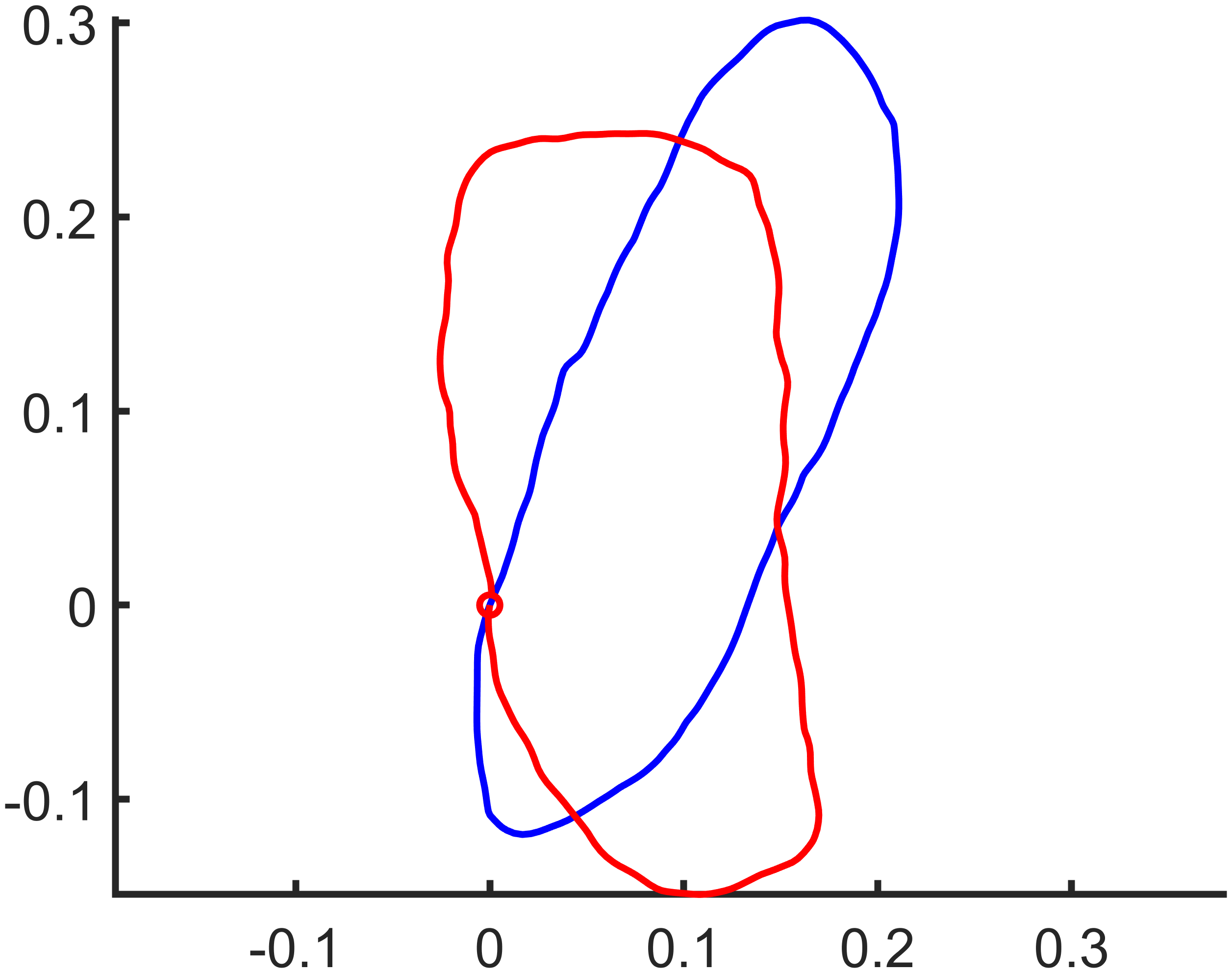}\label{fig:four reparameterizations4}}
	\caption{ {Distance between two cells in the shape space: in blue the contour of the first cell; in red the contour of the second cell, with four parameterizations. The starting point is in the origin ($0,0$). Note that the first shape does not change its initial point, while the second considers four initial points for each evaluated parameterization. The contours are represented and traversed in a counterclockwise direction, starting from the defined initial point on the parameterized shape.}}
	\label{fig:four reparameterizations}
    \end{center}
\end{figure}

{In Figure \ref{fig:four reparameterizations}, an example of the calculation of the distance between two cells in the shape space is shown, considering four reparameterizations of the contour of the second cell. The first cell is represented with the blue outline, and the second is represented with the red outline. For each reparameterization shown, the distance \( d \) is obtained and displayed according to the metric in \( \mathcal{S}_2 \). It can be seen that the second shape (red contour) starts its parameterization each time at a different point on the contour. The starting point of the reparameterization to be analyzed is always at the origin \( (0,0) \). The contours are represented and traversed in a counterclockwise direction, starting from the starting point defined in the parameterized shape.} It is observed that the smallest distance between the shapes is obtained when both have parameterizations corresponding to orientations closer to a total alignment between the shapes along their major axis; in this case, \( d = 0.3586 \).

\begin{figure}[h]
\setlength{\tabcolsep}{0pt} 
	\begin{tabular}{lll} 
			\hskip -0.4 cm\includegraphics[width=4cm]{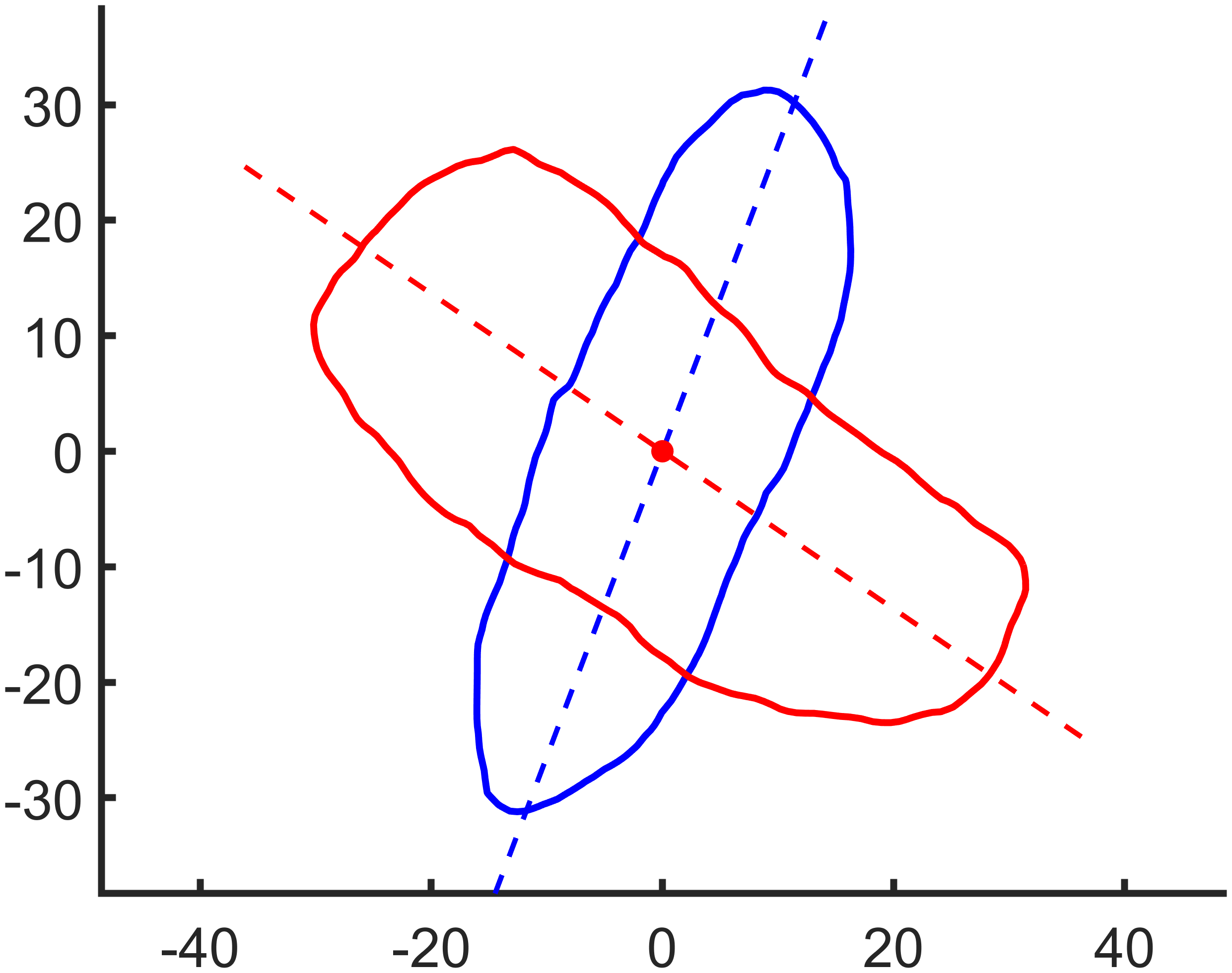}&
            \includegraphics[width=4cm]{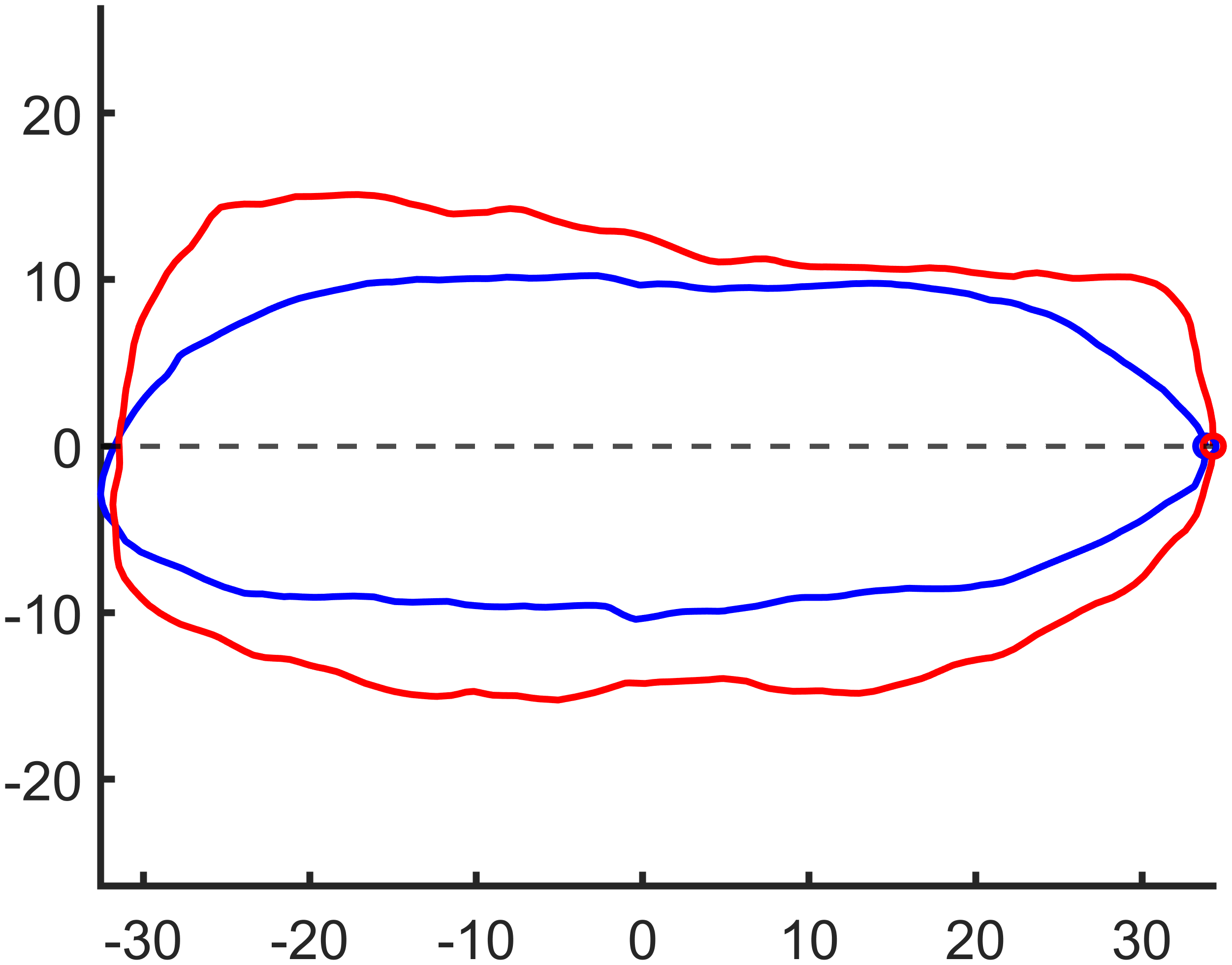}&
            \includegraphics[width=5cm]{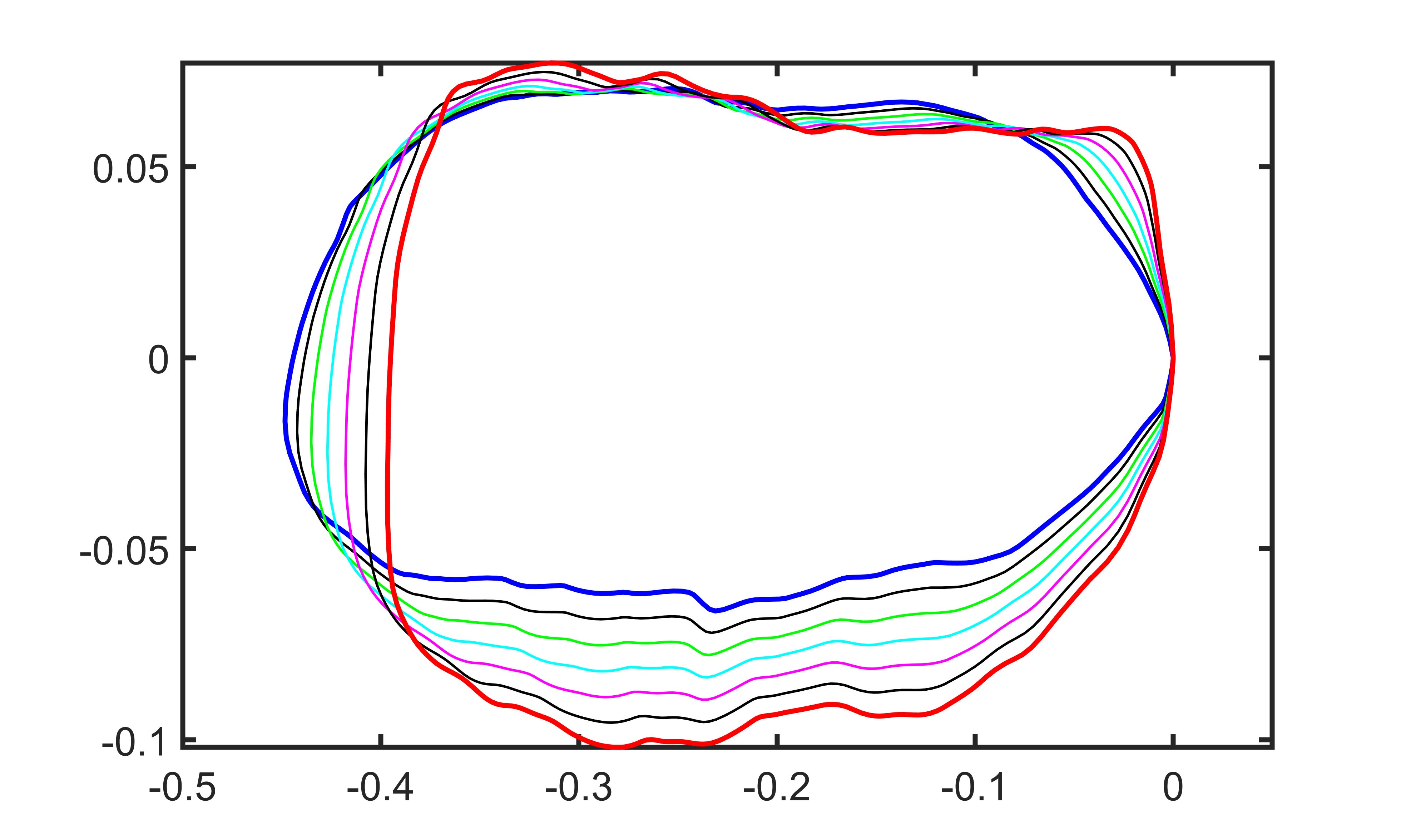}\\	
	\end{tabular}
	\caption{{Distance between the two cells in Figure \ref{fig:four reparameterizations} according to the proposed method. The first two representations correspond to each original contour (blue) and its alignment (red) in the plane. The last one corresponds to the representation of the curves aligned in the shape space (the first in blue, the second in red) and the geodesic that joins them. The distance in this case is $d= 0.2676$.}}
	\label{fig:examples_fixing_parameterization}
\end{figure}

The proposed heuristic considers only one parameterization, which is fixed. {In Figure \ref{fig:examples_fixing_parameterization}, an example of how to apply this strategy is shown. The same cells of Figure \ref{fig:four reparameterizations} are shown, to which the alignment process considered by the proposed method is applied and the initial point of the parameterization is determined, which is located at the intersection of each contour with the $x$ axis on its positive side. This parameterization is fixed, and the distance \textit{d} is obtained according to the metric in $\mathcal{S}_2$: $d= 0.2676$. Here, the computational cost of obtaining the distances between each pair of cells in the space of planar shapes decreases with the previous proposals: in (Figure \ref{fig:four reparameterizations}) reparameterizations of the second shape are performed to determine distances and the smallest of all is considered; with the proposed method (Figure \ref{fig:examples_fixing_parameterization}), a prior alignment of the shapes is conducted to determine a parameterization that is fixed and then obtain the distance, which is considered close to the minimum. In Section \ref{variab_ANOVA}, a statistical study is carried out that allows us to know the variability of the distance obtained using this proposal, in order to validate this fact}.  It is important to note here that when selecting the major axis, the important thing is that this axis will determine the starting point for all arc length settings considered, thus fixing one setting for all curves. Therefore, rotating the cells according to this major axis does not alter the distances but helps to visualize the starting point for all parameterizations.

\section{Experimental Framework}\label{Result_Disc}

We have implemented the techniques described in Section~\ref{section:methods} using MATLAB R2016a 64-bit. The implementation of alignment between shapes is straightforward. {An open source implementation of the general family of elastic metrics, available from \url{https://github.com/h2metrics/h2metrics}, can be used to calculate the distances in both cases. For SRVF, the experiments were performed using SRVF framework toolbox available at \url{http://ssamg.stat.fsu.edu}.} {A version of the implementation of the proposed method is made available at \cite{HeroldGarcia2024}.}


\subsection{Image Database}
\label{subsec:image_database}

We have applied our method to the \emph{erythrocytesIDB} \cite{gonzalez:2015} image database\footnote{Available at \url{http://erythrocytesidb.uib.es}}. These images were taken at the “Dr. Juan Bruno Zayas Alfonso” General Hospital in Santiago de Cuba. { The dataset consists of $202$ images of normal cells, $210$ images of sickle cells, and $211$ images of cells with other cellular deformations. Figure \ref{fig:examples_cell} presents some examples from the image database. The class of erythrocytes with other deformations in this dataset includes cells with minor deformations and shapes resembling normal or sickle cells, which increases the complexity of classifying this category. The database provides predefined contours of the objects, obtained using an active contour method. We employed the contours available in the database, as determining the best segmentation is not the purpose of this work.}

\begin{table}[h]
	\centering
	\resizebox{12cm}{2cm}{
	\begin{tabular}{c } 
		\includegraphics[scale=1]{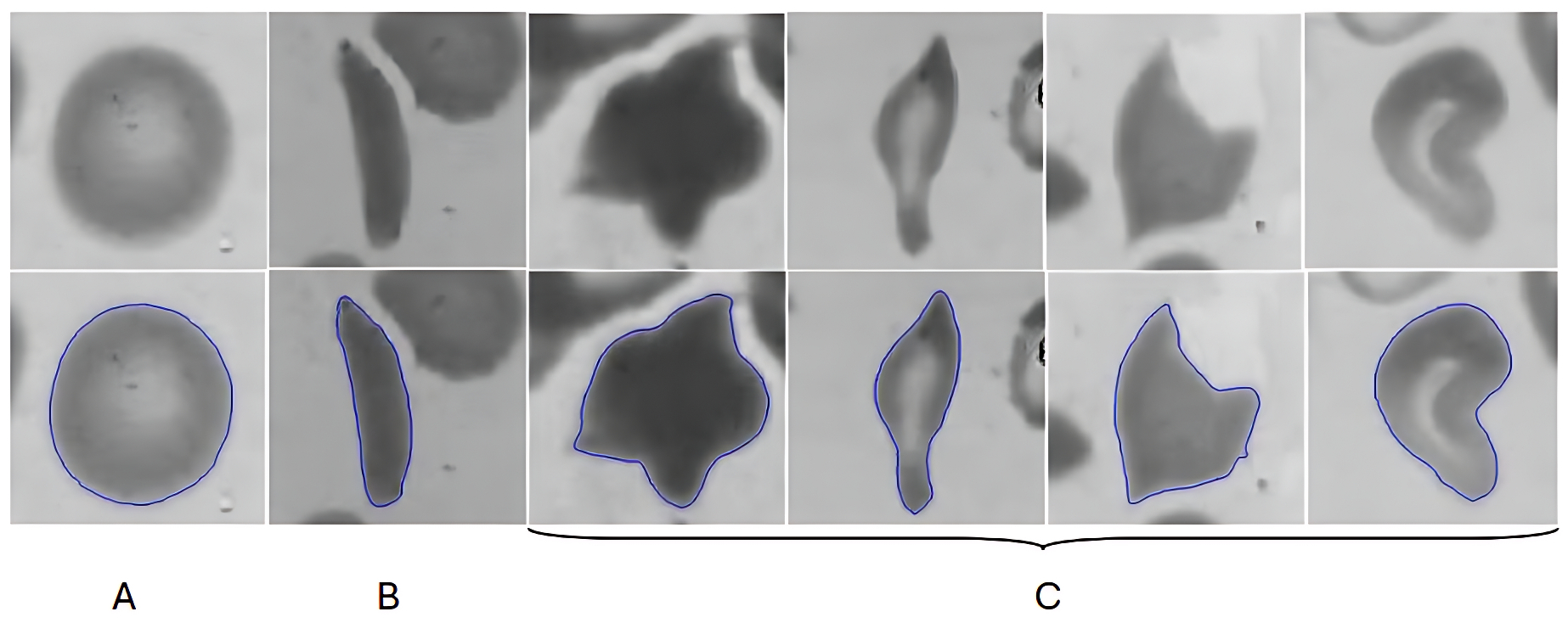}
	\end{tabular}}
	\captionof{figure}{{Examples of erythrocytes classified as Normal~(A), Sickle~(B), and Other Deformations~(C). Up: original cell; Down: the perimeter of segmented regions (in blue).}}
	\label{fig:examples_cell}
\end{table}

{The image dataset used contains objects with shapes close to the Normal or Sickle class within the Other Deformations class, which can be considered cases of input shapes halfway between a circle and an ellipse. It should be noted that supervised classification using this metric in the space of planar shapes provides distances that establish a very pronounced closeness between objects of the Normal class and a circle, and similarly between the Sickle class and an ellipse. Consequently, a shape that is not sufficiently close to these objects is classified as an element of the class Other Deformations. The supervised classification performed demonstrates that the distance metric is able of differentiate the shapes in most cases, yielding results with a high accuracy value.}

\subsection{Experiments and metrics}

In this section, we describe the experiments conducted to evaluate the performance of the methods for computing the distance in the space of planar shapes: the $\mathcal{S}_1$ space (Sec. \ref{sec:s1Space}); the $\mathcal{S}_2$ space (Sec. \ref{sec:s2Space}); and the method for computing distances with fixed parameterization (Sec. \ref{subsec:fixed_paramet}).

The following conditions were considered:

\begin{itemize}
	\item For the experiments conducted in the $\mathcal{S}_1$ shape space, Eq. \ref{minpa} was used, and for the experiments conducted in the $\mathcal{S}_2$ shape space, Eq. \ref{dist_S2} was used.
    
	\item {We identify the contour with the highest number of points and adjust the others to match this resolution, ensuring that the points are relatively equidistant. Since the proposed method significantly reduces the computation time required, the resolution of the contour has a minimal impact on the response time. Therefore, the experiments are performed using the highest available resolution among the contours. This ensures that the calculated distance is closer to the true value and improves the accuracy in the classification process, allowing a better assessment of the method. In this case, the resolution used was 295 points.}
    

	\item {The $k$-NN algorithm \cite{cover:1967} was used for supervised classification}.
	\item {In the classification using circle and ellipse templates, the LDA algorithm \cite{rao:2009} was used}.
	\item {The $k$-medoid algorithm \cite{kaufman:1990} was used for unsupervised classification}.
	\item $5\times 1$-fold cross-validation was applied for supervised classification.
	\item For the experiments involving reparameterizations of the second shape, the value of $t_0$ (Eq. \ref{minpa}) was set to $5$ (for computational complexity reasons), so that each shape was reparameterized $59$ times.
	\item To validate the proposed heuristic, the major axis of the shapes was determined, establishing the starting point for the parameterisations considered. Then, the distance between each pair of shapes was calculated without reparametrizing the second shape.
	\item To examine how the proposed method groups objects other than the normal or sickle shape, we conducted experiments with $3$, $4$, and $5$ groups.
\end{itemize}

{We conducted the following experiments:}

\begin{enumerate}
	\item Experiment 1: Supervised classification of red blood cells considering the fixed parameterization for $\mathcal{S}_1$ and $\mathcal{S}_2$. We compute:
	
	\begin{itemize}
		
		\item Distance between pairs of cells (Cell classification in Table \ref{tab:conf matrix_measures_fixed_param}).
		
		\item Distance between circle and ellipse templates to the shape that represents the cell (Circle and ellipse templates in Table \ref{tab:conf matrix_measures_fixed_param}).
	\end{itemize}
	
	\item Experiment 2: In order to compare the results, experiments were carried out using the {previous methods} raised in \cite{gual:2015} and \cite{epifanio:2020} for $\mathcal{S}_1$ and $\mathcal{S}_2$:
	\begin{itemize}
		\item Supervised classification of red blood cells considering reparameterization of one of the shapes (Cell classification in Table \ref{tab:conf matrix_measures_reparamet}).
		\item Distances between circle and ellipse templates and the cells, with reparameterization of the shapes that represent the cells (Circle and ellipse templates in Table \ref{tab:conf matrix_measures_reparamet}).
	\end{itemize}
	
	\item Experiment 3:  Unsupervised clustering using distance on the shape space considering a fixed parameterization: 
	\begin{itemize}
		\item Two templates (circle and ellipse) and the cells with fixed parameterization. See Table \ref{tab:clustering_FP_S2}.
	\end{itemize}
	\item Experiment 4: Unsupervised clustering classification with previous methods: 
	\begin{itemize}
		\item Reparameterization of one of the shapes. See Table \ref{tab:clustering_FP_S2}.
		\item Two templates (circle and ellipse) and the cells, with reparameterization of the shapes that represent the cells. See Table \ref{tab:clustering_FP_S2}.
		
	\end{itemize}
	
\end{enumerate}

{To evaluate the performance of each of the analyzed experiments, we use the metrics obtained from the confusion matrix, as shown in Table \ref{tab:Classif_ThreeClass}, with the three classes considered in this study: Normal (N), Sickle Cell (S), and Other Deformations (OD).}

{Let {N, S, OD} be the three classes, and the estimated class be denoted by {$\widetilde{N}$, $\widetilde{S}$, $\widetilde{OD}$}, respectively. We denote by $n_{ij}$ ($1 \leq i, j \leq 3$) the number of objects of class $i$ that were classified as belonging to class $j$. The confusion matrix for our problem is:}

\begin{table}[h!]
	\centering
	\begin{tabular}{|c|c|c|c|}
		\hline &N & S & OD\\
		\hline $\widetilde{N}$ & n$_{11}$& n$_{12}$& n$_{13}$\\
		\hline $\widetilde{S}$& n$_{21}$& n$_{22}$& n$_{23}$  \\  
		\hline $\widetilde{OD}$& n$_{31}$& n$_{32}$& n$_{33}$ \\
		\hline
	\end{tabular}
	\caption{{Confusion matrix for classification with three classes}}
	\label{tab:Classif_ThreeClass}
\end{table}

{Being \textit{TP} the number of elements classified as true positives, the performance measures for each class \(i\) are then computed as:}

{Sensitivity (TPR): The number of TP over the total number of objects in that class.}

\begin{equation*}\label{sensitivity}
	TPR=\frac{n_{ii}}{\sum_{j=1}^{3}n_{ij}}.
\end{equation*}

{Precision (P): The number of TP over the total number of instances classified as positive in that class.} 

\begin{equation*}\label{precision}
	P=\frac{n_{ii}}{\sum_{i=1}^{3}n_{ji}}.
\end{equation*}

{F-score (\(F_1\)): The harmonic mean between precision and sensitivity. It provides a balanced measure between both metrics.} 

\begin{equation*}\label{F-score}
	F_i= 2*\frac{P_{i}*TPR_{i}}{P_{i}+TPR_{i}}.
\end{equation*}

{Accuracy (Acc): The proportion of correct predictions made by the model over the total number of predictions.} 

\begin{equation*}\label{accuracy}
	Acc=\frac{\sum_{i=1}^{3}n_{ii}}{\sum_{i=1}^{3}\sum_{j=1}^{3}n_{ij}}.
\end{equation*}

Finally, we also use the SDS-score (SDS): This measure, proposed in \cite{delgado:2020}, allows for the evaluation of the classification to support the study of sickle cell disease, with the three classes analyzed. It is calculated as the quotient of the sum of true positives for the three classes and the number of sickle cells classified as cells with other deformations and vice versa, divided by the sum of the previous numerator and the sum of the incorrect classifications related to normal cells. The SDS-score is intended to serve as an indicator that the results provided by the method are useful to support the analysis of the disease under study.

\begin{equation*}\label{SDS-score}
	SDS= \frac{\sum_{i=1}^{3}n_{ii}+n_{23}+n_{32}}{\sum_{i=1}^{3}\sum_{j=1}^{3}n_{ij}}.
\end{equation*}

This metric takes into account the minimization of false negatives for sickle and other deformed cells, and false positives for normal cells. From a medical point of view, this is of vital importance as it helps avoid a false diagnosis of the patient's good state of health.

\section{Results and Discussion}\label{discussion}
The next section presents and analyzes the results of the experiments conducted. {The best Acc and SDS values obtained in each experiment are highlighted in bold.}

\subsection{{Experiment 1: Supervised Classification Results with Fixed Parameterization on ${\cal S}_1$ and ${\cal S}_2$}}

To carry out the designed experiments, the distances were calculated following our proposal. For the experiment considering distances between cells, the confusion matrix and metrics obtained for the shape spaces ${\cal S}_1$ and ${\cal S}_2$ are shown in the ``Supervised classification" column of Table \ref{tab:conf matrix_measures_fixed_param}.

\begin{table}[h]
	\centering
	\resizebox{12cm}{2cm}{
		\begin{tabular}{|l|c|c|c|c|c|c|c|c|c|c|c|c|}
			\hline \multicolumn{13}{|c|}{Fixed parameterization}\\
			\hline &\multicolumn{6}{|c|}{{Supervised classification}} & \multicolumn{6}{|c|}{Circle and ellipse templates}\\
			\hline Space &\multicolumn{3}{|c|}{${\cal S}_1$}&\multicolumn{3}{|c|}{${\cal S}_2$}&\multicolumn{3}{|c|}{${\cal S}_1$}&\multicolumn{3}{|c|}{${\cal S}_2$}\\
			\hline  & N & S & OD & N & S & OD & N & S & OD & N & S & OD\\
			\hline N & 202 & 0 & 0 & 200 & 2 & 0 & 196 & 0 & 6 & 202 & 0 & 0 \\ 
			\hline S &1 & 199 & 10 & 2 & 197 & 11 & 0 & 209 & 1 & 0 & 194 & 16\\
			\hline OD& 39 & 12 & 160 & 43 & 143 & 25 & 24 & 13 & 174 & 1 & 8 & 202\\
			\hline \multicolumn{7}{|c|}{}&\multicolumn{6}{|c|}{}\\
			\hline TPR & 100 & 94.76 & 75.83 & 99.01 & 93.81 & 11.85 & 97.03 & 99.52 & 82.46 & 100 & 92.38 & 95.73 \\
			\hline P & 83.47 &94.31 & 94.12 & 81.63 & 57.60 & 69.44 & 89.09 & 94.14 & 96.13 & 99.51 & 96.04 & 92.66\\
			\hline $F_1$ & 90.99 & 94.54 & 83.99 & 89.49 & 71.38 & 20.24 & 92.89 & 96.76 & 88.78 & 99.75 & 94.17 & 94.17\\
			\hline Acc / SDS & \multicolumn{3}{|c|}{\textbf{90} / \textbf{93.58}}& \multicolumn{3}{|c|}{68.01 / 92.46} & \multicolumn{3}{|c|}{94 / 95.18} & \multicolumn{3}{|c|}{\textbf{96.03} / \textbf{99.84}} \\
			\hline
	\end{tabular}}
	\caption{Confusion matrix and measures for classification with fixed parameterization in the ${\cal S}_1$ and ${\cal S}_2$ spaces. {(Experiment 1)}.}
	\label{tab:conf matrix_measures_fixed_param}
\end{table}

{The discussion of results considers that each row of the confusion matrix represents elements of class $i$ classified as belonging to class $j$. These results indicate that fixed parameterization provides superior metrics in the ${\cal S}_1$ space. All classes show $F_1$ scores greater than $83\%$. In this space, the classification of the Normal class achieves $F_1 = 90.99\%$. No normal cells were classified as either sickle or other deformation. For the Sickle cell class, $F_1$ is greater than $94\%,$ but there is confusion with the Other Deformations class, which is unfavorable because it is the class of greatest interest. In the Other Deformations class, the $F_1$ value is $83.99\%$, and there is a high degree of confusion with the Normal class and, to a lesser extent, with the Sickle class.} This is due to the similarity of some elements in the Other Deformations class to normal and sickle shapes. We can conclude that the supervised classification with the proposed method is better in the ${\cal S}_1$ shape space, with $Acc = 90\%$ and $SDS = 93.58\%$.

For the experiment considering distances between circle and ellipse templates and cells, the confusion matrix and metrics obtained for the shape spaces ${\cal S}_1$ and ${\cal S}_2$ are shown in the ``Circle and ellipse templates" column of Table \ref{tab:conf matrix_measures_fixed_param}.

The results show that our method is very effective when using circle and ellipse templates in the ${\cal S}_2$ space. {For the Normal class, $F_1 = 99.75\%$. For the Sickle cell class, it is observed that there is confusion when classifying $16$ cells from the Other Deformations class as sickle-shaped, and $F_1 = 94.17\%$}. In the Other Deformations class, the $F_1$ value is $94.17\%$. Here, the confusion between the classes persists in both spaces. In ${\cal S}_1$, $13$ instances of the Other Deformations class are classified as Sickle, and $24$ as Normal, while in ${\cal S}_2$, $8$ instances are classified as Sickle and only one as Normal. In general, the proposed method performs better in the ${\cal S}_2$ shape space, where Acc is 96\% and SDS is 99.84\%, which demonstrates the utility of the proposal in this space for the study of the disease. In this experiment, the results show that with the proposed method there is a {greater tendency to correctly classify the cells}, {and this is because the distance to the circle and ellipse templates effectively discriminates between the three classes, overcoming the challenges associated with classifying based on distances between cells.}

\subsection{Experiment 2: Supervised classification results with {previous methods}}

In order to compare the results, we performed the above experiments using the methods proposed in \cite{gual:2015} and \cite{epifanio:2020}. The confusion matrix and the metrics obtained using the method proposed for the supervised classification of erythrocytes in the shape spaces ${\cal S}_1$ and ${\cal S}_2$ are shown in the ``Supervised classification" column of Table \ref{tab:conf matrix_measures_reparamet}.

\begin{table}[h]
	\centering
	\resizebox{12cm}{2cm}{
		\begin{tabular}{|l|c|c|c|c|c|c|c|c|c|c|c|c|}
			\hline 	\multicolumn{13}{|c|}{Shape reparameterization}\\
			\hline  &\multicolumn{6}{|c|}{{Supervised classification}}&\multicolumn{6}{|c|}{Circle and ellipse templates}\\
			\hline Space &	\multicolumn{3}{|c|}{${\cal S}_1$}&\multicolumn{3}{|c|}{${\cal S}_2$}&\multicolumn{3}{|c|}{${\cal S}_1$}&\multicolumn{3}{|c|}{${\cal S}_2$}\\
			\hline  & N & S & OD & N & S & OD & N & S & OD & N & S & OD\\
			\hline N & 202 & 0 & 0 & 202 & 0 & 0 & 198 & 0 & 4 & 199 & 0 & 3 \\ 
			\hline S & 17 & 186 & 7 & 0 & 206 & 4 & 0 & 197 & 13 & 0 & 201 & 9\\
			\hline OD & 49 & 11 & 151 & 15 & 7 & 189 & 20 & 17 & 174 & 11 & 13 &187\\
			\hline \multicolumn{7}{|c|}{}&\multicolumn{6}{|c|}{}\\
			\hline TPR & 100 & 88.57 & 71.56 & 100 & 98.10 & 89.57 & 98.02 & 93.81 & 82.46 & 98.51 & 95.71 &88.63\\
			\hline P & 75.37 & 94.42 & 95.57 & 93.09 & 96.71 & 97.93 & 90.83 & 92.06 & 91.10 & 94.76 & 93.93 &93.97\\
			\hline $F_1$ & 85.96 & 91.40 & 81.84 & 96.42 & 97.40 & 93.56 & 94.29 & 92.92 & 86.57 & 96.60 & 94.81 &91.22\\
			\hline Acc / SDS & \multicolumn{3}{|c|}{87 / 89.41} & \multicolumn{3}{|c|}{\textbf{96.03} / \textbf{97.59}} &\multicolumn{3}{|c|}{92 / 96.15} & \multicolumn{3}{|c|}{\textbf{94.02} / \textbf{97.75}} \\
			\hline
	\end{tabular}}
	\caption{Confusion matrix and measures for the classification with reparameterization in the ${\cal S}_1$ and ${\cal S}_2$ spaces. {(Experiment 2)}.}
	\label{tab:conf matrix_measures_reparamet}
\end{table}

{The highest Acc of 96.03\% and an SDS of 97.59\% were achieved in the ${\cal S}_2$ shape space.} {It is observed that the Normal class achieves $F_1 = 96.42\%$. In the Sickle cell class, $F_1 = 97.40\%$, and for the Other Deformations class, $F_1 = 93.56\%$. It can be observed that although in both spaces there is confusion when classifying objects of the Sickle and Other Deformations classes, the errors increase with fixed reparameterization for supervised classification (see Table \ref{tab:conf matrix_measures_fixed_param} ``Supervised classification"). This is the expected behavior, since obtaining the distance between cells provides more information in the case of previous methods, and therefore the values obtained are more representative and capable of discriminating between classes. However, these values do not exceed those achieved in the case of supervised classification with fixed parameterization using distances to circle and ellipse templates (see Table \ref{tab:conf matrix_measures_fixed_param} ``Circle and ellipse templates"), where $Acc=96.03\%$ and $SDS=99.84\%$. This is because, when calculating distances to two templates, this value is less affected by the variability of the cell contours, which only occurs in one of the shapes that participate in the distance calculation.}

The confusion matrix and metrics obtained for the classification process by computing the distance between circle and ellipse templates with reparameterization of the shape that represents the cell in ${\cal S}_1$ and ${\cal S}_2$ are shown in the ``Circle and ellipse templates" column of Table \ref{tab:conf matrix_measures_reparamet}.

{In this case, the highest Acc (94.02\%) and SDS (97.75\%) values in space ${\cal S}_2$ are lower than those achieved with the proposed method in the same space, which were 96.03\% and 99.84\%, respectively (Table \ref{tab:conf matrix_measures_fixed_param}, ``Circle and ellipse templates'').
An analysis of the values achieved in each class shows that for the Normal class, $F_1 = 96.60\%$,
and for Other Deformations class, $F_1 = 91.22\%$. These values do not exceed those obtained in space ${\cal S}_2$ with the proposed method, which were $99.75\%$ and $94.17\%$, respectively (Table \ref{tab:conf matrix_measures_fixed_param}, ``Circles and ellipse templates"). For the Sickle class, the $F_1$-score of 94.81\% is slightly higher than the 94.17\% achieved in the ${\cal S}_2$ space with the proposed method, and both demonstrate a performance exceeding 94\%.
This highlights the effectiveness of the proposed method, particularly for the Sickle and Other Deformations classes, which are the ones of greatest interest.} The best SDS is obtained in the ${\cal S}_2$ space with the proposed method $(99.84\%)$, which validates the usefulness of our method in the study and treatment of sickle cell disease.

\subsection{Experiment 3: Unsupervised clustering results considering fixed parameterization in ${\cal S}_2$}

The distance between planar shapes has also been previously used in an unsupervised clustering algorithm to define homogeneous class deformations \cite{gual:2015, epifanio:2020}. In this section, we also use the distance between the planar shapes presented in Sec. \ref{sec:s2Space} as a measure of similarity in an unsupervised clustering algorithm, considering the proposed heuristic. {These experiments were conducted exclusively in the ${\cal S}_2$ space using template distances, as this approach achieved better results in the supervised classification experiments. We use the $k$-medoid algorithm, which selects $k$ points from the dataset as medoids,  and then assigns each point in the dataset to the nearest medoid, ensuring that the differences between points within each group are minimized.}




{Initially, clustering is performed with $k = 3$, since the problem involves analyzing three classes: Normal, Sickle, and Other Deformations. For this value of $k$, the ``Fixed Parameterization" column in Table \ref{tab:clustering_FP_S2} presents the confusion matrix obtained by calculating the distances between the circle and ellipse templates and the shapes, considering their fixed parameterization.}

The results show that our method is very effective in the classification of the {Normal class, with $F_1$ (99.75\%). The Sickle cell class achieved an $F_1$ score of 93.86\%, while Other Deformations scored 94.01\%}. In general, the method reached $Acc = 96\%$ and $SDS = 100\%,$ demonstrating its superior performance in accurately grouping similar shapes for the study of sickle cell disease, as it consistently classifies the classes correctly.

\subsection{Experiment 4: Unsupervised clustering results with {previous methods}}

To compare the results, we performed the experiments using the methods proposed in \cite{epifanio:2020}. {The ``With reparameterization" column in Table \ref{tab:clustering_FP_S2} shows the confusion matrix  obtained using distances between shapes, taking into account the reparameterization of one of the shapes, and the use of the distance from circle and ellipse templates with reparameterization of the shape.}

\begin{table}[h]
	\centering
	\resizebox{12cm}{2cm}{
		\begin{tabular}{|l|c|c|c|c|c|c|c|c|c|}
			\hline \multicolumn{10}{|c|}{Unsupervised clustering}\\
			\hline &\multicolumn{3}{|r|}{Fixed parameterization}& \multicolumn{6}{|c|}{With reparameterization }\\
			\hline ${\cal S}_2$ &\multicolumn{3}{|c|}{Circle and ellipse templates} & \multicolumn{3}{|c|}{Cell clustering}&\multicolumn{3}{|c|}{Circle and ellipse templates}\\
			\hline  & N& S& OD& N & S & OD & N & S & OD \\
			\hline N &202&0&0& 200 & 0 & 2 & 200 & 0 & 2\\ 
			\hline S &0&191&19& 0 & 200 & 10 & 0 & 202 & 8\\
			\hline OD &1& 6&204& 37 & 23 & 151 & 17 & 6 & 188\\
			\hline \multicolumn{4}{|c|}{}&\multicolumn{6}{|c|}{}\\
			\hline TPR &100&90.95&96.68& 99.01& 95.24 & 71.56 & 99.01 &96.19 &89.10\\
			\hline P &99.51&96.95&91.48& 84.39 & 89.69 & 92.64 & 92.17 & 97.12 & 94.95\\
			\hline $F_1$ &99.75&93.86&94.01& 91.12 & 92.38 & 80.75 & 95.47 & 96.65 & 91.93 \\
			\hline Acc / SDS & \multicolumn{3}{|c|}{\textbf{96 / 100}}&\multicolumn{3}{|c|}{89 / 94} & \multicolumn{3}{|c|}{\textbf{95} / \textbf{97}} \\
			\hline
	\end{tabular}}
	\caption{Confusion matrix and measures of shape classification for unsupervised clustering for $k=3$ in ${\cal S}_2$ space {(Experiments $3$ and $4$)}.}
	\label{tab:clustering_FP_S2}
\end{table}

{The best results for Acc and SDS in this case are obtained using distances to circle and ellipse templates, with $95\%$ and $97\%$, respectively}. These values do not exceed those achieved using the proposed method and distances to templates, which are $96\%$ and $100\%$, respectively. {In general, our method tended to correctly classify the cells}. {Crucially, no sickle cells were misclassified as normal, ensuring accuracy in critical cases.}

To test how this framework can cluster objects that do not correspond to the normal or sickle shape, which are our classes of most interest, we performed experiments using our method with groups of $4$ and $5$ cells. {The Average Silhouette Width (ASW) metric (see \cite{batool2021}) was used to determine the quality of the clusters (see Table \ref{tab: Group_Generated}).}

\begin{table}[ht]
	\centering
	\begin{tabular}{|l|c|c|c|c|c|c|c|c|c|c|}
		\hline ${\cal S}_2$& N& S& G1& G2& G3& N& S& G1& G2& G3\\
		\hline &\multicolumn{5}{|c|}{$k = 4$ (ASW = $0.75$)}&\multicolumn{5}{|c|}{$k = 5$ (ASW = $0.64$)}\\
		\hline N &136 &2 &1 &72 &-&135& 71& 1& 4& 0\\
		\hline S& 11& 188& 0& 11& -&6& 11& 0& 89& 104\\
		\hline OD& 0& 0& 202& 0& -&0& 0& 202& 0& 0\\
		\hline
	\end{tabular}
	\caption{Unsupervised clustering in ${\cal S}_2$ for groups generated with $k=4$ and $k=5$}
	\label{tab: Group_Generated}
\end{table}

{Table \ref{tab: Group_Generated} shows the results of the experiments performed for $k = 4$ and $k = 5$, specifying the ASW values: $0.75$, $0.64$, respectively. For higher values of $k$, the ASW was as follows: $0.62$ for $k=6$; $0.73$ for $k=7$; and $0.70$ for $k=8$. In all cases, the ASW closest to $1$ is obtained for $k=4$, which shows that in this case, on average, better grouping is achieved}.

{The experiment confirmed that high quality clusters can be achieved using the $ \mathcal{S}_2$ space, maintaining quality in the classes of greatest interest for the study of the disease: normal and sickle cells. However, it is important to note that this is a preliminary experiment, as the images used represent exclusively sickle cell blood samples, which exhibit a higher prevalence of the morphological deformations associated with sickle cell disease.}


 \subsection{Computational cost}

{The previous approaches to obtain the morphological classification of erythrocytes in the shape spaces ${\cal S}_1$ and ${\cal S}_2$, need an iterative process to obtain the minimum distance between the shapes, expressed by Eq. (\ref{minpa}) and Eq. (\ref{dist_S2}). The actions carried out within the iterative process are considered to be of constant order $C$, because they are the same in all cases. Furthermore, the computational cost and the amount of memory required are determined by the number of reparameterizations $t_0$ studied and the number $k$ of cells analyzed.} 


{In the worst case, $t_0 = N$, where $N$ represents the number of points of the analyzed contour. Considering $k$ cells to analyze, the computational cost of obtaining the distance between an initial curve and the remaining $k - 1$ shapes is of order $O(C\cdot N\cdot k)$.}

{Since $C$ is constant, the order is considered to be $O(N\cdot k)$. Therefore, to obtain the distance between each pair of shapes representing the $k$ cells analyzed, the computational cost will be $O(N\cdot k^2)$.}
{The proposed heuristic uses a fixed parameterization, setting $N = 1$, which reduces computational complexity. Thus, the computational cost of obtaining the distances between each pair of cells in the space of planar shapes is of order $O(k^2)$. In the case of the analysis performed using the distance between the shapes representing the cells and the circle or ellipse templates, the computational cost decreases to $O(2k)$, since it is only necessary to calculate distances to two known shapes, so it is considered to be of order $O(k)$.}

\subsection{Distance variability}\label{variab_ANOVA}


{This research proposes aligning the cells based on their major axis to obtain a distance between shapes that approaches the minimum. This approach enables us to obtain robust practical results in classification, while eliminating the iterative process required by previous methods. To verify that the distances obtained for each parameterization are statistically consistent across the two classes of greatest interest (normal and sickle cells), an analysis of variance was conducted.}

We took a sample of $50$ instances of the Sickle class and $50$ instances of the Normal class. From each instance, we obtained  distance measures for four different parameterizations. Then we calculate the average distance of each group for each of the established parameterizations. 

Table \ref{tab:ANOVA} shows that for the parameterization factor (row), {the critical value of $F$ ($2.818$) is greater than the value of $F$( $0.210$),  indicating that parameterization does not significantly affect the dependent variable distance. Similarly, for the class factor (column), $F$ ($4.844$) exceeds $F$ ($0.001$), confirming that class also has no significant effect on the dependent variable distance}. The assumption of homogeneity of the variances between the parameterizations and classes is tested and it is { confirmed that the experimental samples have equal variance within the analyzed population.}

\begin{table}[h!]
	\centering
	\resizebox{12cm}{1cm}{
	\begin{tabular}{|l|c|c|c|c|c|c|}
		\hline Origin variations&Sum squares&Degrees freedom&Square mean& F& Probability&Critical value F\\
		\hline Rows &0.064 &11 &0.006 &0.210 &0.992&2.818\\
		\hline Columns& 0& 1& 0& 0.001&0.972&4.844\\
		\hline Mistake& 0.307& 11& 0.028& -&-&-\\
		\hline Total& 0.372& 23& -& -& -&-\\
		\hline
	\end{tabular}}
	\caption{Analysis of variance results.}
	\label{tab:ANOVA}
\end{table}

{\subsection{Accuracy comparison with state-of-the-art models}\label{comparison}}

{When comparing our classification results with those obtained by other approaches, each method must applied to the same dataset. For this reason, all studies referenced in this section utilize the erythrocytesIDB dataset (see Subsection \ref{subsec:image_database}).} 
\medskip

{In \cite{gual:2015}, the space of planar shapes \( \mathcal{S}_1 \) is considered, employing a distance invariant only under arc-length parameterizations. The classical \( k \)-Nearest Neighbor ($k$-NN) algorithm with contour descriptors derived from integral-geometry methods is applied in \cite{gualetal:2015}. Three standard supervised classifiers (Naïve Bayes, $k$-NN, and Support Vector Machine (SVM)) based on a set of nine numerical shape features are explored in \cite{rodrigues2016morphological}. Scenario 1 of \cite{alzubaidi2020deep} utilizes a deep learning model directly with original images as input. In \cite{PETROVIC2020104027}, the classification method and features yielding the best performance for cell morphology analysis are selected. The space of planar shapes \( \mathcal{S}_2 \), along with the elastic metric defined by the square-root velocity function, is used in \cite{epifanio:2020}. Finally, \cite{gualvaya23} involves estimating two contour shape parameters for classification using stereological methods.}
\medskip

{The highest accuracy levels achieved by each method are presented in Table \ref{resultadosOtrosMetodos}. All compared methods demonstrate high classification accuracies, exceeding 91\%. Among these, our proposed methodology stands out with one of the highest performance percentages.}
\medskip

{Beyond accuracy comparisons, we evaluated the computational costs of methods in \cite{gual:2015}, \cite{gualetal:2015}, and \cite{epifanio:2020}. While two of these approaches achieve accuracies close to 96\%, their computational costs are substantially higher than those of our proposed method, highlighting its efficiency.}

\begin{table}[htb]
\begin{center}
\begin{tabular}{|c|c|}
\hline
 Method & Accuracy (\%) \\
 \hline
 Gual-Arnau {\it et al.} (2015). {\it Image Anal Stereol.} \cite{gual:2015} & 93\\
  Gual-Arnau {\it et al.} (2015). {\it Med. Biol. Eng. Comput.} \cite{gualetal:2015} & 96\\
 Rodrigues {\it et al.} (2016) {\it Workshop de Visao Computacional.} \cite{rodrigues2016morphological}& 95 \\
De Faria {\it et al.} (2018) {\it Workshop de Visao Computacional.} \cite{defaria:2018}& 94 \\
Alzubaidi {\it et al.} (2020) {\it Electronics.} \cite{alzubaidi2020deep}& 91 \\
Petrović {\it et al.} (2020) {\it Comput. Biol. Med.} \cite{PETROVIC2020104027}& 95 \\
Epifanio {\it et al.} (2020) {\it  Image Anal Stereol.} \cite{epifanio:2020}& 96 \\
Gual-Vayà (2024) {\it  Image Anal Stereol.} \cite{gualvaya23}& 93 \\
 Our method & 96 \\
 \hline

\hline
\end{tabular}
\caption{{Comparison of accuracy results from previous methods on the erythrocytesIDB dataset with our proposed method}.} \label{resultadosOtrosMetodos}
\end{center}
\end{table}
\bigskip

{\subsection{Application of the proposed method to other shape analysis scenarios}\label{applications}}

{The proposed method demonstrates efficiency and accuracy for clinical applications in sickle cell disease analysis, and it shows potential for extension to other geometric shape analysis tasks. As an example, its application is demonstrated on several objects from the well-known Flavia plant leaf dataset, one of the first significant datasets for leaf classification introduced by Wu et al. \cite{Wu2007}.}

\begin{figure}[!ht]
    \begin{center}
    \includegraphics[width=2.5cm]{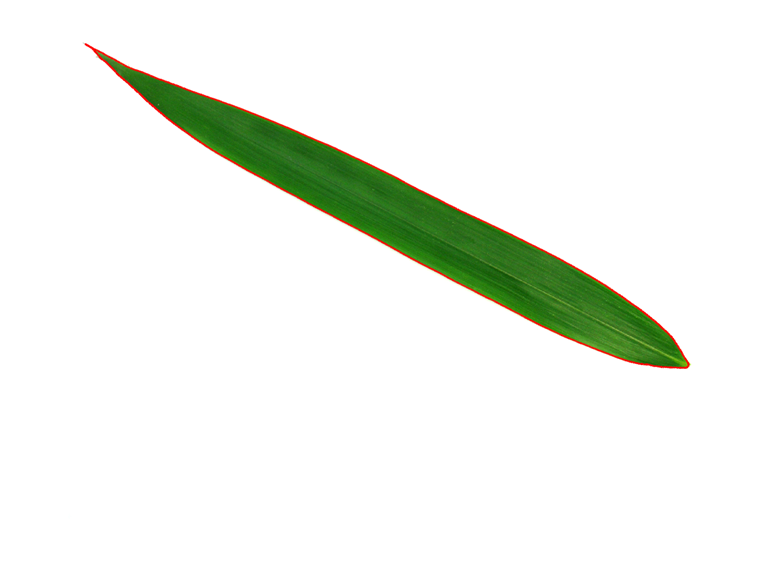}\quad
    \includegraphics[width=2.5cm]{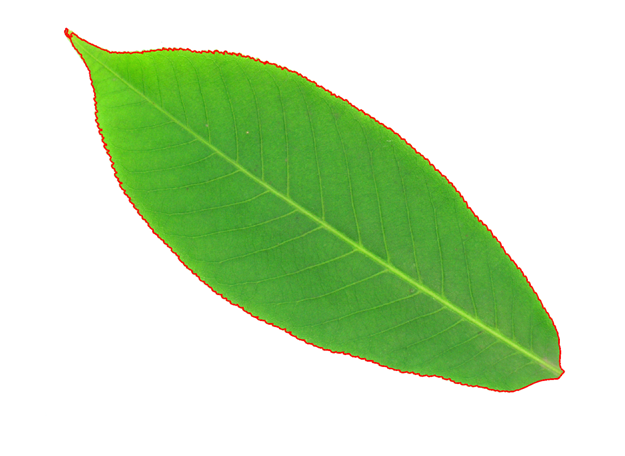}\quad
    \includegraphics[width=2.5cm]{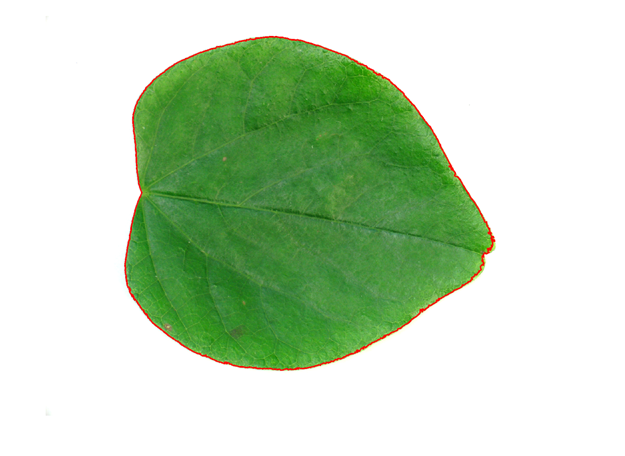}\quad
    \includegraphics[width=2.5cm]{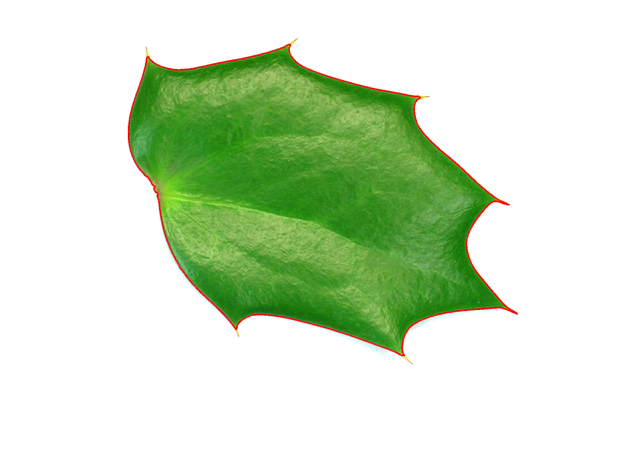}\\
    \includegraphics[width=2.5cm]{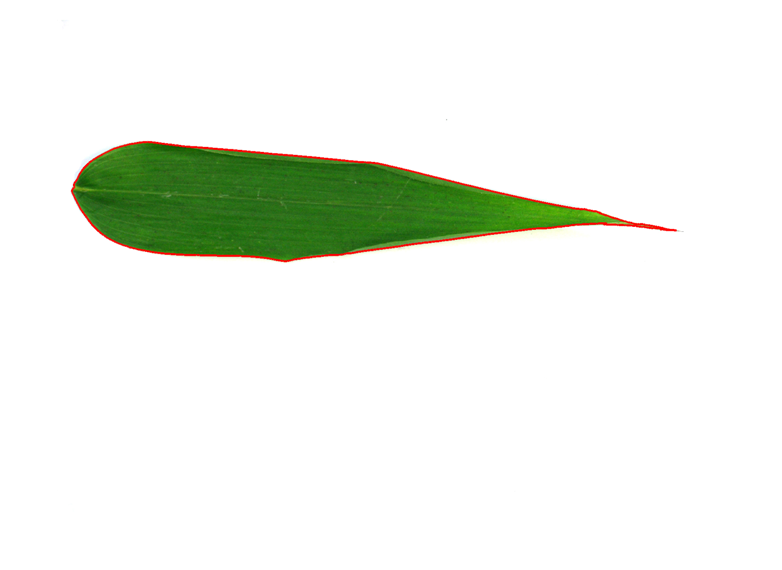}\quad
    \includegraphics[width=2.5cm]{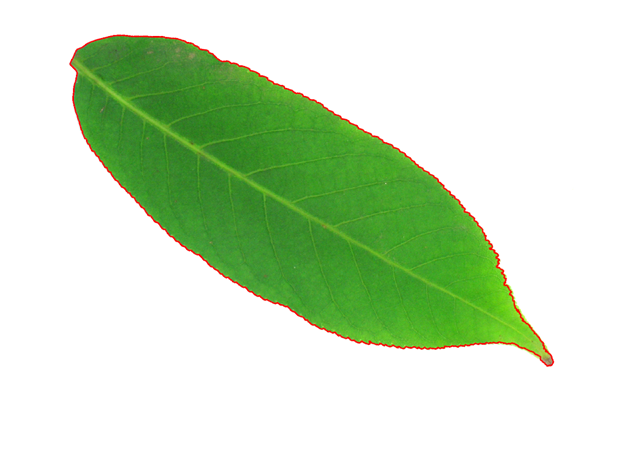}\quad
    \includegraphics[width=2.5cm]{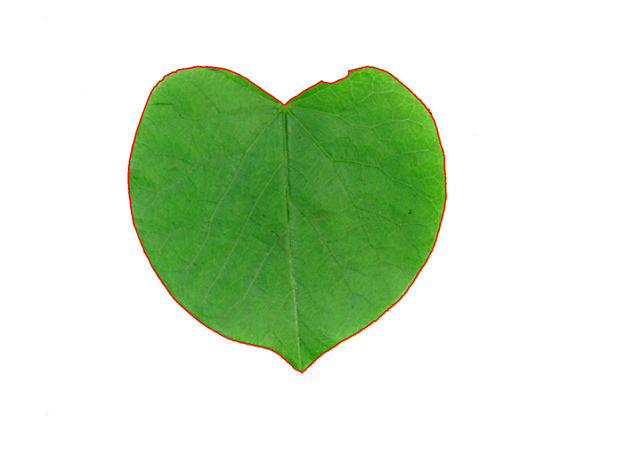}\quad
    \includegraphics[width=2.5cm]{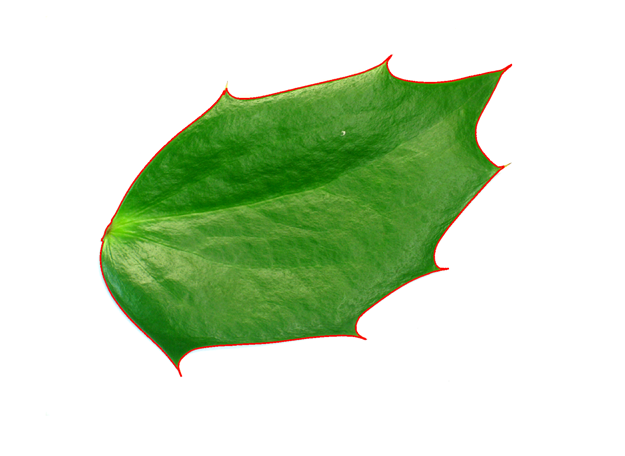}
    \caption{{Examples of the Flavia data set. In columns, from left to right, two samples of each type of leaves. Contours detected in red.}}
    \label{fig:leaves}
    \end{center}
\end{figure}

{Two samples from four different leaf types proposed by this dataset were considered. Figure \ref{fig:leaves} shows the leaves and their detected contours (in red) used for processing.}

{Table \ref{tab:leaves_distances} presents the geodesic distances between objects in the Flavia dataset, calculated using the proposed method. The results show that the distances between objects of the same class are consistently smaller compared to the distances between objects of different classes.}

\begin{table}[h!]
	\centering
	\begin{tabular}{|l|c|c|c|c|c|c|}
		\hline Classes&Type 1&Type 2&Type 3& Type 4\\
		\hline Type 1 & 0.3473 & 0.5553 & 0.6651 & 0.5965 \\
		\hline Type 2 &      & 0.5127 & 0.6758 & 0.7489 \\
		\hline Type 3 &      &      & 0.6371 & 0.7371\\
		\hline Type 4 &      &      &      & 0.5935 \\
		\hline
	\end{tabular}
	\caption{Distances between objects of dataset Flavia.}
	\label{tab:leaves_distances}
\end{table}

\begin{figure}[!ht]
    \begin{center}
    \subfloat[Type 1 - Type 1]{\includegraphics[width=5cm]{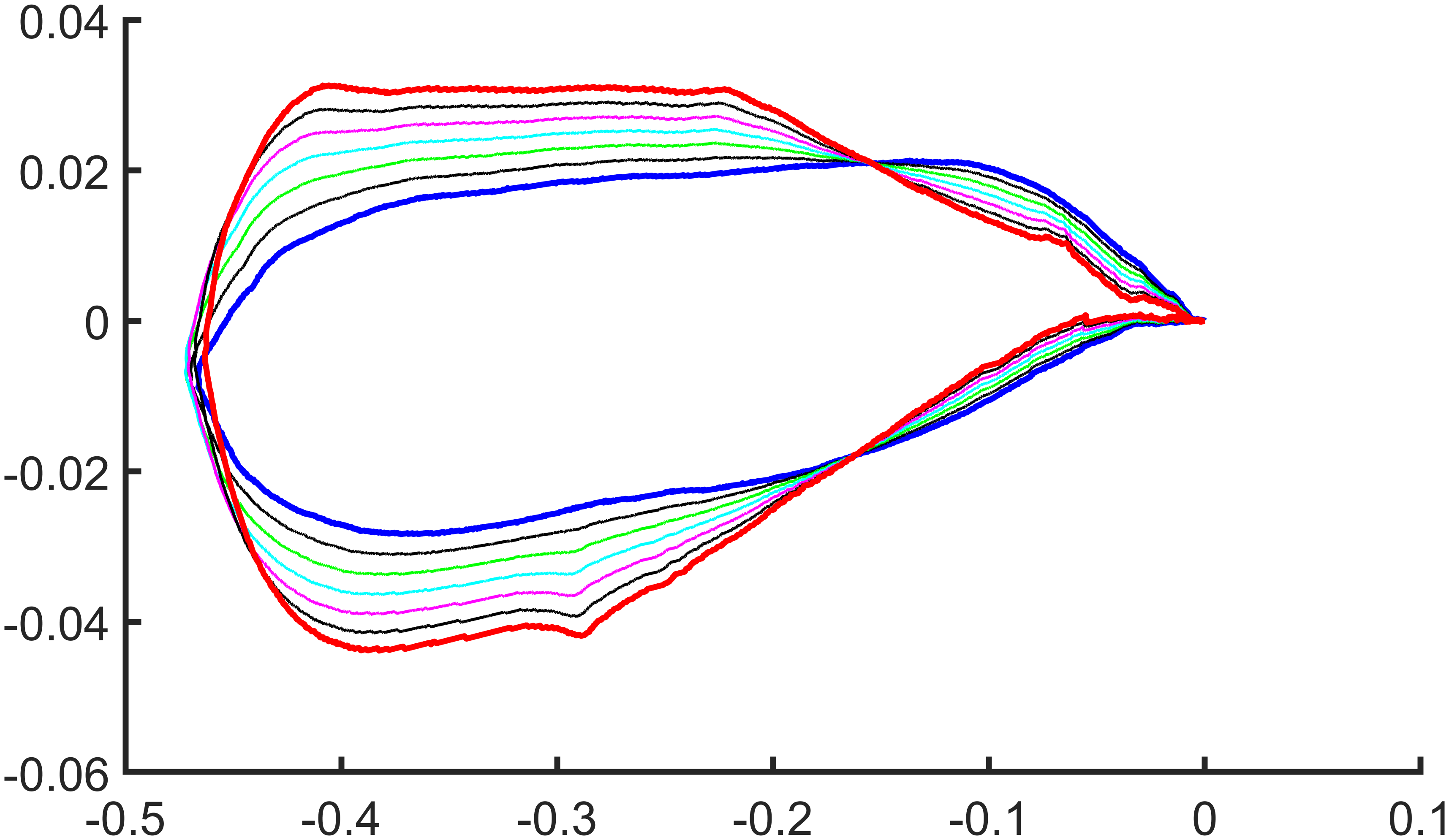}}\quad
    \subfloat[Type 1 - Type 2]{\includegraphics[width=5cm]{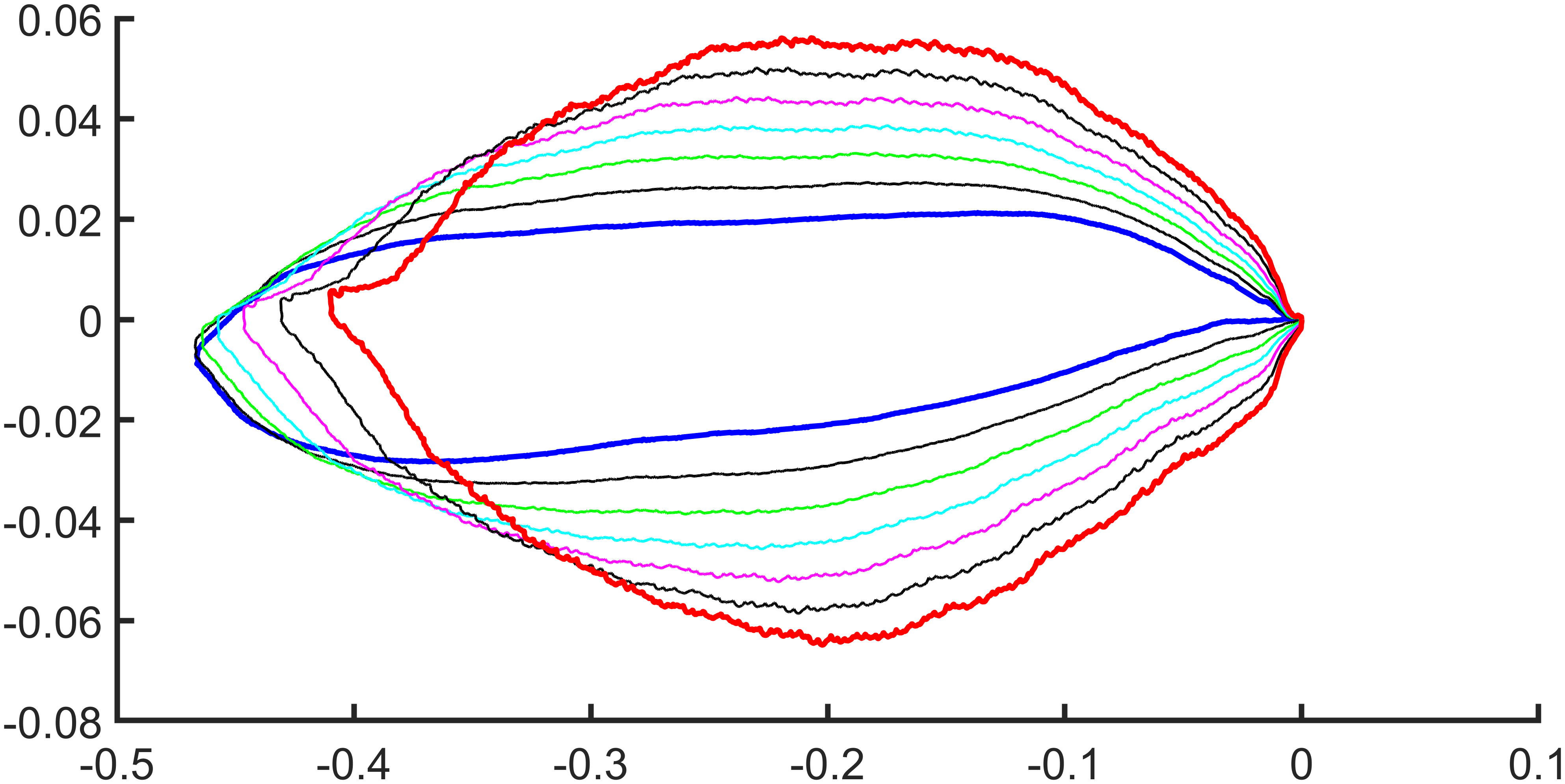}}\\
    \subfloat[Type 3 - Type 3]{\includegraphics[width=5cm]{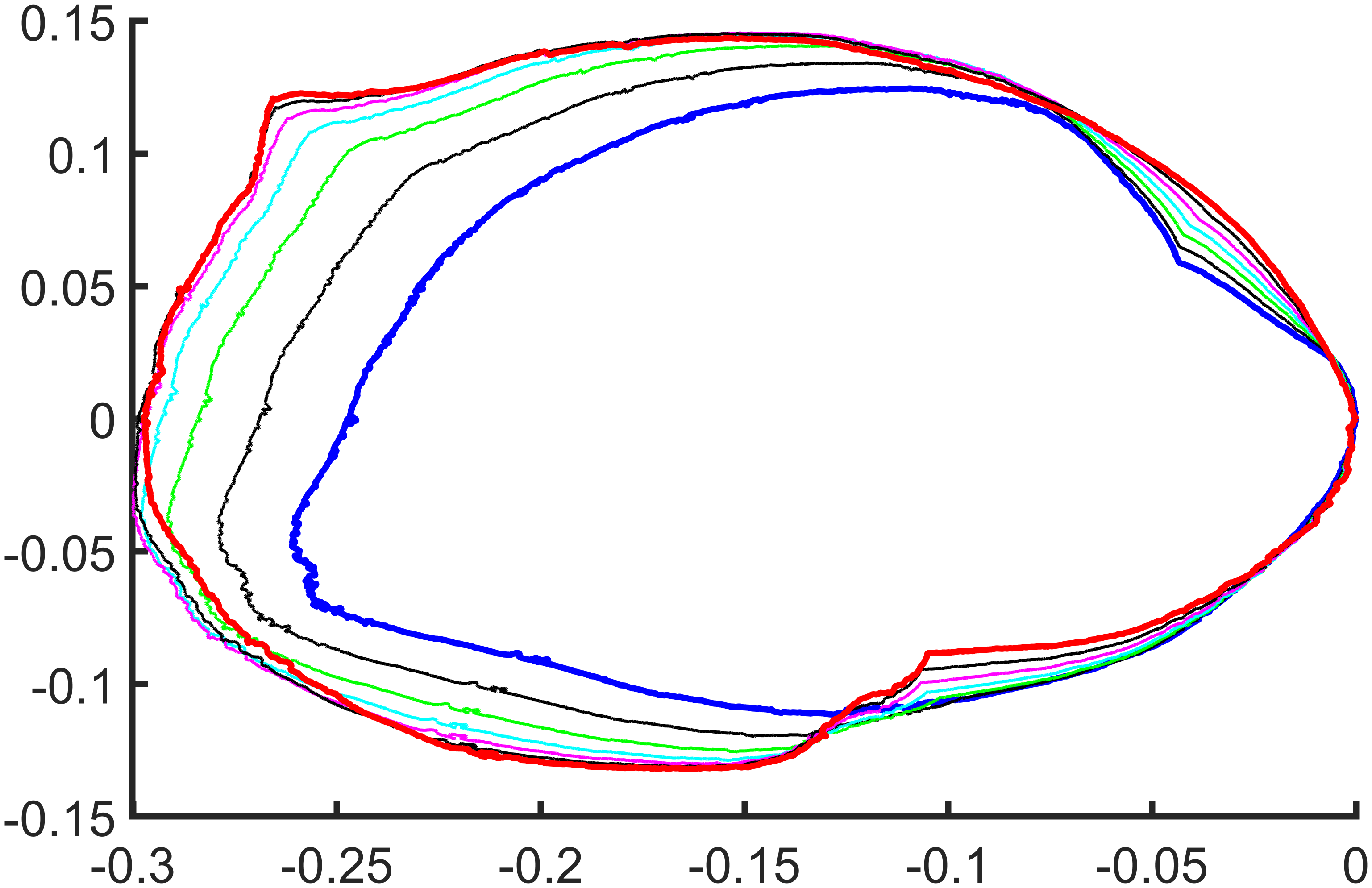}}\quad
    \subfloat[Type 3 - Type 4]{\includegraphics[width=5cm]{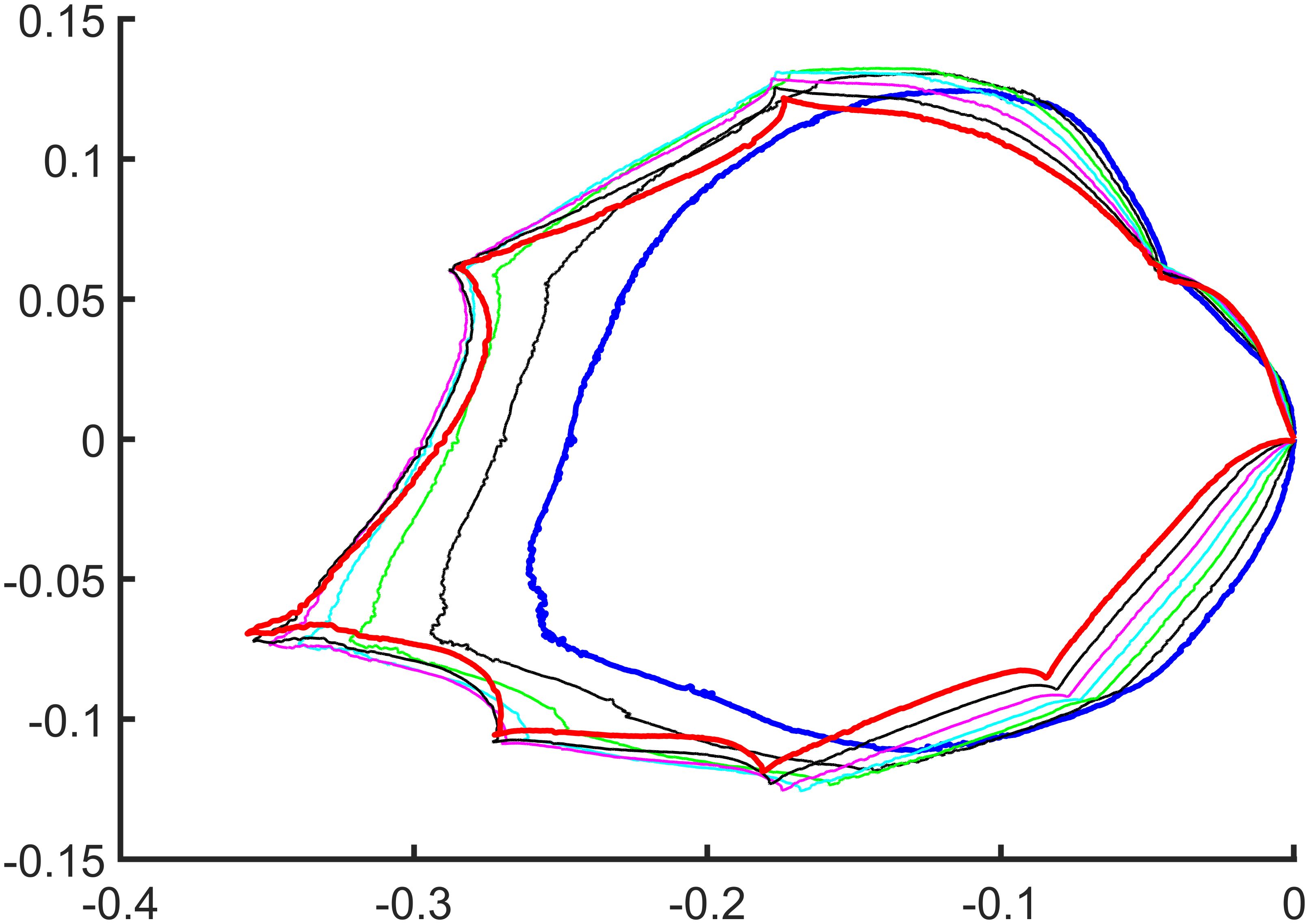}}
    \caption{{Geodesic trajectories obtained in some examples of the classes considered from the Flavia data set.}}
    \label{fig:leaves_geod}
    \end{center}
\end{figure}

{Some examples of the geodesic paths obtained in the evaluated cases are illustrated in Figure \ref{fig:leaves_geod}.}

{\subsection{Considerations about the proposal}\label{considerations}}

{The proposed method demonstrates efficiency and high accuracy in classifying sickle cells, effectively avoiding false negatives. Its use of a balanced image dataset further ensures minimal bias across classes. The reduction in computational cost positively impacts the efficiency of obtaining results and makes its application feasible in environments with limited computational resources or in real-time analysis systems. This gives the model significant potential for practical application, particularly in tools that assist specialists in analyzing large volumes of images. It is well-suited for clinical applications in the treatment of sickle cell disease and has the potential to be extended to other geometric shape analysis scenarios. Additionally, the approach of considering contours as closed curves in the shape space is innovative and provides a foundation for future studies exploring cellular deformation processes. However, it has some limitations that are important to consider when evaluating its possible use in shape classification processes.}

{The main limitation is the dependence on the geometric templates used, which were chosen due to the similarity that the erythrocytes characteristic of sickle cell anemia have with circles and ellipses. In fact, other pathologies cause different erythrocyte deformations, and although unsupervised experiments were carried out to consider this possibility that achieved highly accurate results, this limitation will probably reduce the efficiency of the method's performance if it is used to classify other structures with contours that differ greatly from these templates, as occurs in cases of contours that follow shapes with a high degree of non-convexity. This same limitation must be considered in cases of generalization of the method to other shape analysis environments, such as the example shown in Section \ref{applications}}.

{The second limitation is related to the quality of the contours to be studied. Contour resolution can be affected by various factors, such as noise in image capture, which can be introduced by situations related to the quality of the equipment used and the preparation of the sample, or the performance of the segmentation methods used.}

{Finally, the image database used is limited to images from a single health center (see Section \ref{subsec:image_database}), so the characteristics inherent to these images are specific to the population of the region where this center is located, and may not be representative of other populations with different genetic, environmental or health characteristics. In addition, the images are specific to a pathology, sickle cell anemia, so they do not have representations of other types of cellular deformations corresponding to other pathologies. Future work could involve expanding the dataset to include diverse pathologies, improving the model's robustness and generalizability. Additionally, before integrating the method into tools for medical diagnostic support, it is essential to conduct preliminary experiments to validate its performance in these new scenarios.}

\bigskip

\section{Conclusions}\label{conclu}

{In this study, we address the challenge of automatically analyzing erythrocyte shapes in peripheral blood smear images of sickle cell disease to classify them as normal cells, sickle-shaped cells, or cells with other deformations. Effective classification methods require balancing high accuracy with low computational costs. A proven approach represents cell shapes as planar curves based on their contours, using well-defined metrics to enable classification algorithms. However, these methods are computationally expensive as they require minimizing distances across all parameterizations of closed curves.}

{We propose a novel heuristic approach for cell classification by leveraging the characteristic shapes of healthy and sickle cells through two complementary ideas. First, we define a fixed parameterization based on each cell's major axis. Second, we calculate distances to two templates: a circle and an ellipse. This strategy eliminates the need to minimize distances across all parameterizations, reducing computational cost from $O(N\cdot k^2)$ to $O(N\cdot k)$.}


{We conducted four experiments to evaluate our method, comparing it with previous solutions based on shape space analysis ([33], [2]). The model’s effectiveness was assessed using a confusion matrix with indicators such as Accuracy, Precision, Sensitivity, $F_1$-score, and SDS. The results show an Acc = 96.03\%, outperforming previous approaches (91.02\% and 94.02\%). The $F_1$-scores were 99.75\% for Normal and 94.17\% for both Sickle and Other Deformations classes. Additionally, the SDS metric reached 99.84\%, demonstrating the method’s utility in studying sickle cell disease. Clustering experiments further achieved Acc = 96\% and SDS = 100\%  in scenarios with three groups.}

{The proposed approach is efficient, achieving high accuracy in classifying sickle cells while avoiding false negatives and significantly reducing computational cost, making it suitable for clinical applications in sickle cell disease treatment and feasible in resource-limited regions or real-time analysis settings. In addition, this approach in the space of planar shapes opens new avenues for researching cellular deformation processes. The defined distances enable geodesics, facilitating shape interpolation modeling and laying a foundation for further studies in this field.}

{One of the limitations of the method is its dependence on circles and ellipses  templates, which may affect its results when used to analyse other objects that do not meet these conditions, as well as its sensitivity to the quality of the contour. However, if the user takes this into account, they can adapt them to their target shapes.}



{Finally, we emphasize that the methodology presented is interpretable and explainable \cite{arrieta:2020}, featuring simulatability, where a human observer can cognitively follow its decision-making process, and decomposability, allowing each component to be independently understood. Additionally, it offers algorithmic transparency, enabling users to comprehend the procedures generating outputs from input data, a critical feature for fostering trust in healthcare applications.}

\section*{Declaration of Generative AI and AI-assisted technologies in the writing process}
During the preparation of this work, the author(s) used ChatGPT in order to check the correctness of the English in some paragraphs. After using this tool/service, the authors reviewed and edited the content as needed and take full responsibility for the content of the publication.

\section*{CRediT authorship contribution statement}
\textbf{Yaima Paz Soto: }Investigation, Writing – original draft. \textbf{Silena Herold-Garcia: }Conceptualization, Investigation, Methodology, Software, Supervision, Writing – original draft, Writing – review and editing. \textbf{Ximo Gual Arnau: }Conceptualization, Formal analysis, Funding acquisition, Supervision, Writing – original draft, Writing – review and editing. \textbf{Antoni Jaume-i-Capó: }Funding acquisition, Supervision, Writing – original draft, Writing – review and editing. \textbf{Manuel González-Hidalgo: }Funding acquisition, Supervision, Writing – original draft, Writing – review and editing.

\section*{Declaration of competing interest}
The authors have no conflict of interest to declare.

\section*{Funding}

This work was partially supported by: R+D+i Project PID2019-104829RA-I00 “EXPLainable Artificial INtelligence systems for health and well-beING (EXPLAINING)” funded by MCIN/AEI/10.13039/501100011033; Project PID 2020-113870GB-I00 ``Desarrollo de herramientas de Soft Computing para la Ayuda al Diagnóstico Clínico y a la Gestión de Emergencias (HESOCODICE)'', funded by MCIN/AEI/10.13039/501100011033;  Project PID2023-149079OB-I00 funded by MICIU/AEI, Spain/10.13039/5011000 11033/ and ERDF, EU; Project OCDS-REC2023/07\- ``Aproximacions a l'anàlisi morfològica de glòbuls vermells aplicant intel·ligència artificial en l'espai forma com a suport al diagnòstic i seguiment de la drepanocitosi'', funded by Govern de les Illes Balears and Universitat de les Illes Balears; Project PN223LH010-082 ``Imaging Diagnosis of Cellular Deformation in Patients with Sickle Cell Disease in the Eastern Region of Cuba'', funded by the National Program of Basic and Natural Sciences of Cuba; and by the Spanish Ministry of Science and Innovation (PID2020-115930GA-100 and PID2022-141699NB-I00).


\bibliographystyle{elsarticle-num}
\bibliography{Refer.bib}

\end{document}